\documentclass{article}

\usepackage{microtype}
\usepackage{graphicx}
\usepackage{subcaption}
\usepackage{booktabs} % for professional tables
\usepackage{multirow}
\usepackage[table]{xcolor}
\usepackage{float} 
\usepackage{hyperref}
\usepackage{tcolorbox}

\usepackage[preprint]{icml2026}

\usepackage{amsmath}
\usepackage{amssymb}
\usepackage{mathtools}
\usepackage{amsthm}
\usepackage{makecell}
\definecolor{colorours}{HTML}{FFFFE0}
\definecolor{colorhead}{HTML}{e2ecda}
\usepackage[capitalize,noabbrev]{cleveref}

\theoremstyle{plain}

\theoremstyle{definition}

\theoremstyle{remark}

\usepackage[textsize=tiny]{todonotes}

\icmltitlerunning{AdaTSQ: Pushing the Pareto Frontier of Diffusion Transformers via Temporal-Sensitivity Quantization}

\begin{document}

\twocolumn[
  \icmltitle{AdaTSQ: Pushing the Pareto Frontier of Diffusion Transformers via Temporal-Sensitivity Quantization}



  \icmlsetsymbol{equal}{*}
  \vspace{-4mm}
  \begin{icmlauthorlist}
    \icmlauthor{Shaoqiu Zhang}{equal,sjtu}
    \icmlauthor{Zizhong Ding}{equal,sjtu}
    \icmlauthor{Kaicheng Yang}{sjtu}
    \icmlauthor{Junyi Wu}{sjtu}
    \icmlauthor{Xianglong Yan}{sjtu}
    \icmlauthor{Xi Li}{mt}
    \icmlauthor{Bingnan Duan}{mt}
    \icmlauthor{Jianping Fang}{mt}
    \icmlauthor{Yulun Zhang$^{\dagger}$}{sjtu}
  \end{icmlauthorlist}

  \icmlaffiliation{sjtu}{Shanghai Jiaotong University}
  \icmlaffiliation{mt}{Meituan, China}

  \icmlcorrespondingauthor{Yulun Zhang}{yulun100@gmail.com}

  \icmlkeywords{Machine Learning, ICML}
      \begin{center}
        \centering
        \includegraphics[width=0.94\textwidth]{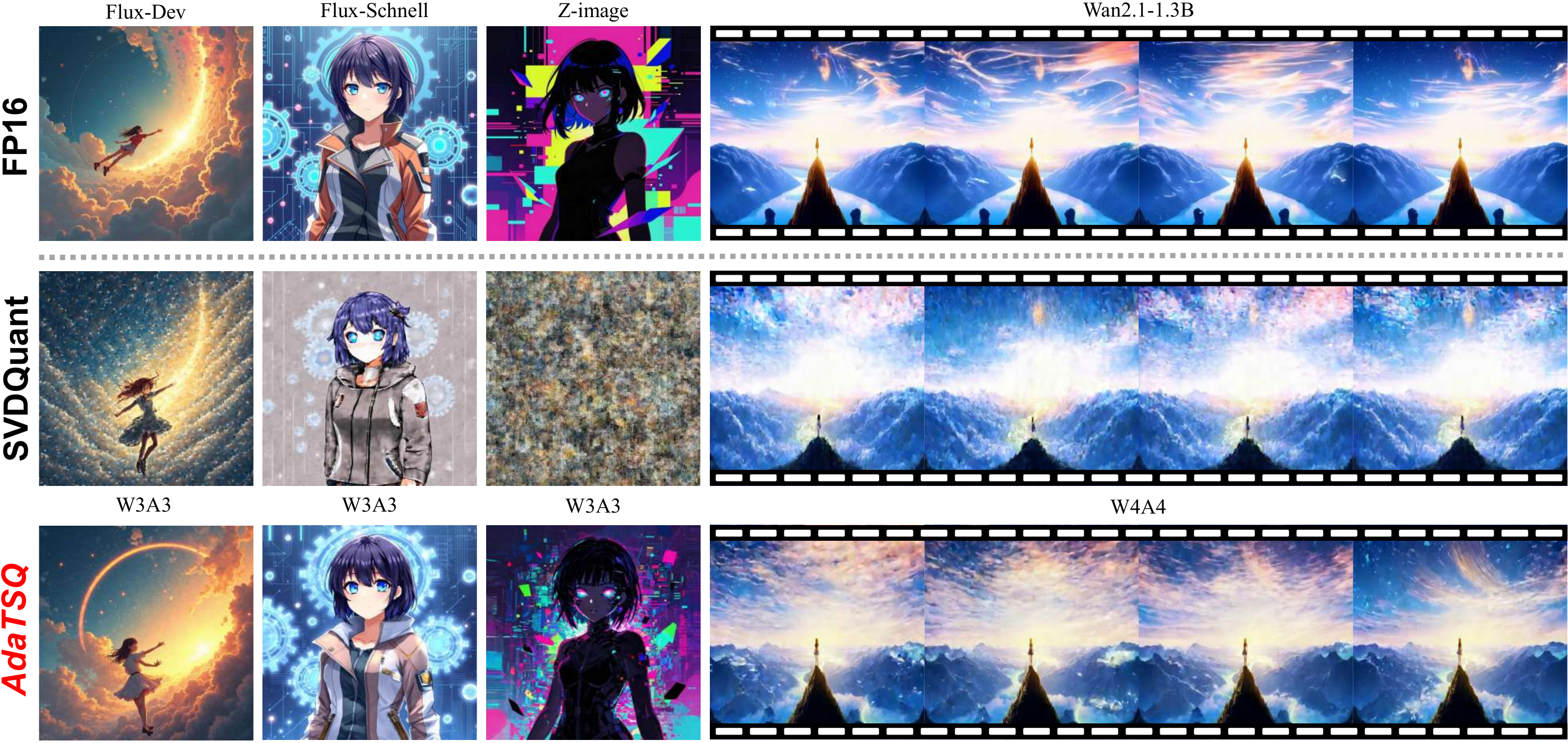}
        \vspace{-2mm}
        \captionof{figure}{Visual comparison of AdaTSQ with FP16 and SVDQuant~\cite{li2024svdquant} under different low-bit quantization settings. The comparison includes three text-to-image models (Flux-Dev, Flux-Schnell, Z-Image) and one text-to-video model (Wan2.1-1.3B).}
        \label{fig:main-comparation}
    \end{center}
  \vskip 0.1in
]



\printAffiliationsAndNotice{\icmlEqualContribution}  % no special notice (required even if empty)

\begin{abstract}
\vspace{-3mm}

Diffusion Transformers (DiTs) have emerged as the state-of-the-art backbone for high-fidelity image and video generation. However, their massive computational cost and memory footprint hinder deployment on edge devices. While post-training quantization (PTQ) has proven effective for large language models (LLMs), directly applying existing methods to DiTs yields suboptimal results due to the neglect of the unique temporal dynamics inherent in diffusion processes. In this paper, we propose \textbf{AdaTSQ}, a novel PTQ framework that pushes the Pareto frontier of efficiency and quality by exploiting the temporal sensitivity of DiTs.
First, we propose a Pareto-aware timestep-dynamic bit-width allocation strategy. We model the quantization policy search as a constrained pathfinding problem. We utilize a beam search algorithm guided by end-to-end reconstruction error to dynamically assign layer-wise bit-widths across different timesteps.
Second, we propose a Fisher-guided temporal calibration mechanism. It leverages temporal Fisher information to prioritize calibration data from highly sensitive timesteps, seamlessly integrating with Hessian-based weight optimization.
Extensive experiments on four advanced DiTs (\textit{e.g.}, Flux-Dev, Flux-Schnell, Z-Image, and Wan2.1) demonstrate that AdaTSQ significantly outperforms state-of-the-art methods like SVDQuant and ViDiT-Q. Our code will be released at \url{https://github.com/Qiushao-E/AdaTSQ}. %Notably, we achieve \textbf{lossless W4A4} quantization across all models and, for the first time, demonstrate \textbf{lossless W3A3} generation on Flow Matching-based Flux models. Our code will be released soon.
\vspace{-10mm}
\end{abstract}

\setlength{\abovedisplayskip}{2pt}
\setlength{\belowdisplayskip}{2pt}
\begin{figure*}[t!]
    \centering
    % Ensure you export your final PPT/Illustrator drawing as a high-res PDF or PNG
    % e.g., figures/overview.pdf
    \includegraphics[width=\textwidth]{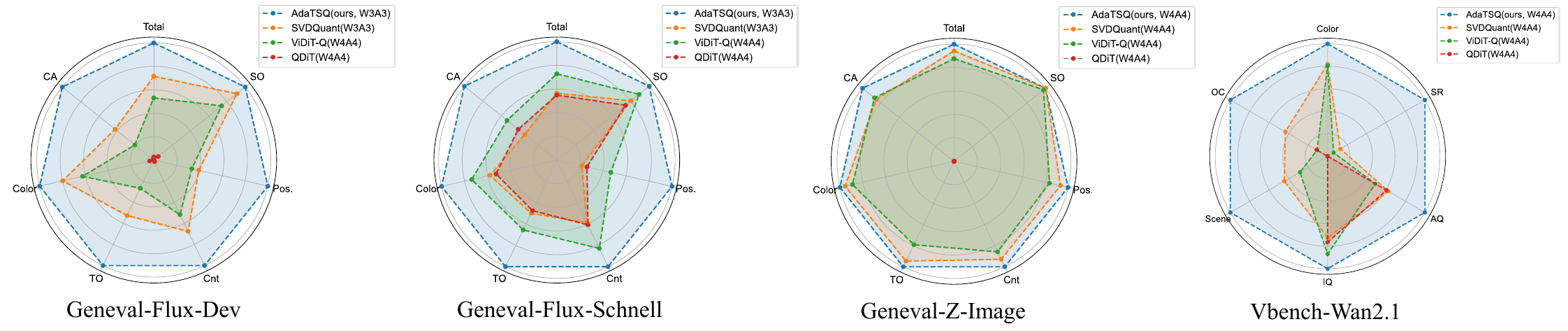}
    \vspace{-4mm}
    \caption{Holistic Performance Comparison. Radar charts comparing AdaTSQ (Blue) with baselines across four DiT models. Axes denote normalized metrics for fidelity, alignment, and consistency. AdaTSQ consistently achieves the largest coverage area, indicating superior comprehensive performance across all modalities and sampling schedules.}
    \label{fig:result-compare}
    \vspace{-5mm}
\end{figure*}

\section{Introduction}
\vspace{-2mm}
The landscape of generative AI has been fundamentally reshaped by Diffusion Transformers (DiTs) \cite{peebles2023scalablediffusionmodelstransformers, Peebles2022ScalableDM, bao2023all, gao2023masked, ma2024sit, lu2024fit, crowson2024scalable}, which have supplanted traditional U-Net architectures \cite{Rombach2022HighResolutionIS, Saharia2022PhotorealisticTD,wu2025flashedit} to establish new state-of-the-art (SOTA) benchmarks in high-fidelity image and video synthesis. Representative models, such as Flux \cite{labs2025flux1kontextflowmatching}, Z-Image \cite{team2025zimage, jiang2025distribution, liu2025decoupled}, and Wan2.1 \cite{wan2025}, demonstrate unparalleled capabilities in generating coherent, photorealistic content. However, this performance comes with substantial computational overhead. Unlike single-pass language models, DiTs require iterative denoising timesteps to genarate videos or images, and when combined with the massive parameter count and the quadratic complexity of self-attention mechanisms, the inference cost becomes prohibitive. This immense memory footprint and latency impede their deployment on resource-constrained edge devices and severely limit their applicability in real-time scenarios.

To mitigate these costs, Post-Training Quantization (PTQ) offers a lightweight alternative to expensive Quantization-Aware Training (QAT). In Large Language Models (LLMs), methods like GPTQ \cite{frantar-gptq}, AWQ \cite{lin2023awq}, FlatQuant \cite{sun2024flatquant}, QuaRot \cite{ashkboos2024quarot}, and DuQuant \cite{lin2024duquant} have successfully achieved 4-bit quantization by handling activation outliers. However, these methods focus on static or token-wise distributions. This raises a pivotal question: \textit{Can these text-centric recipes be directly transferred to the temporally dynamic Diffusion Transformers?}

\begin{figure*}[t!]
    \centering
    % Ensure you export your final PPT/Illustrator drawing as a high-res PDF or PNG
    % e.g., figures/overview.pdf
    \includegraphics[width=\textwidth]{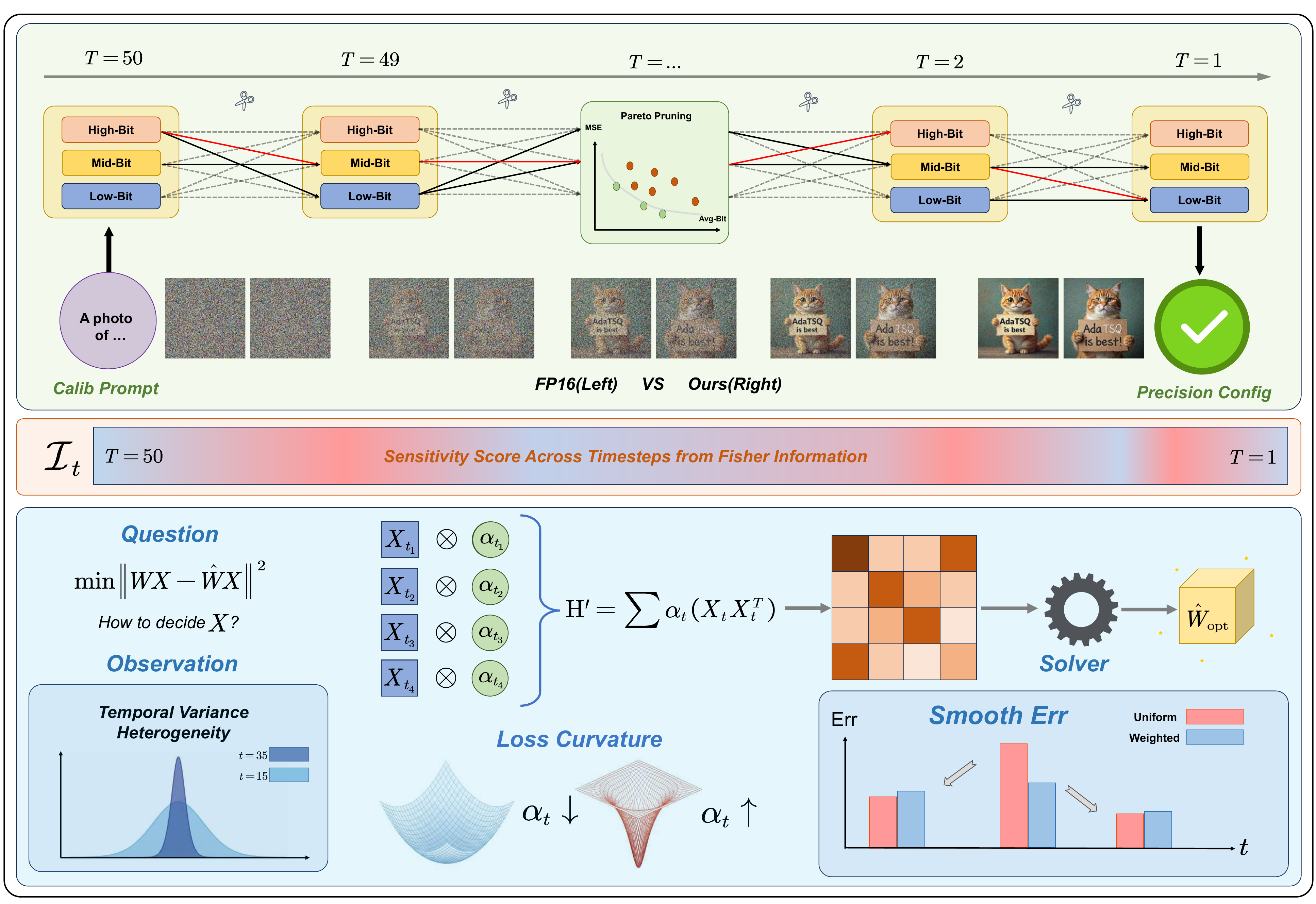}
    \vspace{-6mm}
    \caption{Overview of the \textbf{AdaTSQ} framework. The upper panel illustrates the \textbf{Pareto-aware Timestep-Dynamic Allocation}, which employs beam search to find the optimal bit-width schedule. The lower panel depicts the \textbf{Fisher-Guided Temporal Calibration}, which leverages temporal sensitivity to re-weight the Hessian for risk-aware weight optimization.}
    \label{fig:overview}
    \vspace{-4mm}
\end{figure*}

Unfortunately, the answer is negative. Directly applying LLM methods degrades quality by overlooking the unique temporal dynamics of diffusion. Recent DiT-specific attempts also fall short. ViDiT-Q \cite{zhao2024viditq} uses rotation but relies on suboptimal static Min-Max weight quantization. SVDQuant \cite{li2024svdquant} employs low-rank decomposition but treats all timesteps uniformly, ignoring varying sensitivities. Similarly, S\textsuperscript{2}Q-VDiT \cite{feng2025s} and others \cite{liu2025clq,li2025dvd,yang2025robuq} neglect the critical temporal dimension.

To address these limitations, we argue that the quantization paradigm for DiTs must evolve from static to temporally dynamic. Our motivation stems from two fundamental observations illustrated in Figure \ref{fig:motivation}. First, as shown in the Fisher heatmap (Fig. \ref{fig:motivation}a), the sensitivity of DiT layers to quantization noise is highly non-uniform, exhibiting distinct phases where early steps govern high-level structure and later steps refine details. Second, the activation distributions shift drastically across the temporal axis (Fig. \ref{fig:motivation}b). These findings imply that a uniform bit-width assignment is inherently suboptimal, as it wastes bit-budget on insensitive denoising steps while starving critical ones, and fails to adapt to the evolving activation statistics.

\vspace{-1mm}
Based on this finding, we propose AdaTSQ, a unified framework that harmonizes temporal dynamics with quantization efficiency. First, to tackle the combinatorial complexity of assigning bit-widths across varying timesteps, we propose \textbf{Pareto-aware timestep-dynamic allocation strategy}. Instead of relying on computationally expensive ILP solvers or manual heuristics, we formulate the allocation as a constrained pathfinding problem. By employing an efficient beam search, we navigate the trade-off space to identify the Pareto-optimal bit-width trajectory, minimizing reconstruction error while strictly adhering to average bit-rate constraints. Second, for weight quantization, we move beyond uniform data calibration by proposing the \textbf{Fisher-guided calibration
mechanism}. We derive temporal importance scores from Fisher Information to dynamically re-weight the calibration data. This mechanism ensures that the Hessian-based optimization \cite{frantar-gptq} explicitly prioritizes minimizing errors in the most sensitive timesteps, effectively steering quantization noise away from the critical path of the diffusion process.

\vspace{-2mm}
In summary, our contributions are:
\begin{itemize}
    \vspace{-4mm}
    \item We propose \textbf{AdaTSQ}, a novel PTQ framework that integrates Pareto-aware dynamic allocation strategy and Fisher-guided calibration to exploit the temporal heterogeneity of DiTs' Quantization.
    \vspace{-2mm}
    \item We propose a Pareto-aware timestep-dynamic allocation strategy. It solve the constrained pathfinding formulation via beam search to optimally distribute activation bit-budgets across timesteps
    \vspace{-2mm}
    \item We develop a Fisher-guided temporal calibration strategy to re-weight data, ensuring Hessian-based solvers prioritize accuracy in critical denoising steps.
    \vspace{-2mm}
    \item Extensive experiments demonstrate the effectiveness of AdaTSQ on Flux, Wan2.1, and Z-Image (Figure \ref{fig:result-compare}). We achieve robust W3A3 generation on image models and high-fidelity W4A4 performance on video models.
\end{itemize}
\vskip -0.05in

\section{Related Works}
\vspace{-1mm}
\subsection{Diffusion Transformers}
\vspace{-1mm}
Diffusion Probabilistic Models (DPMs) \cite{ho2020denoising, song2021denoising} have emerged as the dominant framework for high-fidelity generative tasks. Early approaches, such as Latent Diffusion Models (LDMs) \cite{rombach2022high}, typically relied on convolutional U-Net architectures. Recently, the field has decisively shifted towards Diffusion Transformers (DiTs) \cite{peebles2023scalablediffusionmodelstransformers, Peebles2022ScalableDM, bao2023all}. By adopting a Transformer backbone, DiTs leverage superior scalability and global attention mechanisms to handle complex multi-modal data. This architecture now underpins leading text-to-image and text-to-video models, including Wan2.1 \cite{wan2025}, Z-image \cite{team2025zimage,jiang2025distribution,liu2025decoupled}, and Flux \cite{labs2025flux1kontextflowmatching}. Notably, Flux employs Flow Matching \cite{lipman2023flowmatchinggenerativemodeling}, which simplifies the generative process with straighter ODE trajectories and offers potential advantages in training stability. However, despite these improvements, the massive parameter counts and the inherently iterative nature of DiTs result in prohibitive computational costs. This creates an urgent need for efficient compression techniques.

\vspace{-2mm}

\subsection{Quantization}
Post-Training Quantization (PTQ) has become the standard for compressing large-scale models. In the LLM domain, methods like GPTQ \cite{frantar-gptq}, AWQ \cite{lin2023awq}, SmoothQuant \cite{xiao2023smoothquant}, QuaRot \cite{ashkboos2024quarot}, FlatQuant \cite{sun2024flatquant}, and DuQuant \cite{lin2024duquant} successfully pushing the limits to 4 bits. However, transferring these recipes to DiTs is non-trivial. Unlike LLMs, DiTs exhibit drastic temporal shifts in activation distributions \cite{zhao2024viditq}. Early diffusion quantization focused on U-Nets \cite{li2023qdiffusion, shang2023ptqdm}, but attention has recently shifted to DiT-specific challenges \cite{chen2024QDiT, wu2024ptq4dit}. Advanced methods like ViDiT-Q \cite{zhao2024viditq}, SVDQuant \cite{li2024svdquant}, S\textsuperscript{2}Q-VDiT \cite{feng2025s}, and others \cite{li2025dvd, wu2025quantcache, liu2025clq} achieve 4-bit quantization via rotation or decomposition. Yet, a critical gap remains. These methods enforce uniform bit-widths or rely on heuristic rules, failing to exploit the varying sensitivity of different denoising phases. In contrast, our work propose a theoretically grounded, Pareto-optimal strategy to allocate bit-budgets across the temporal dimension.

\begin{figure*}[t]
    \centering
    \begin{subfigure}[b]{0.4\textwidth}
        \centering
        \includegraphics[width=\textwidth]{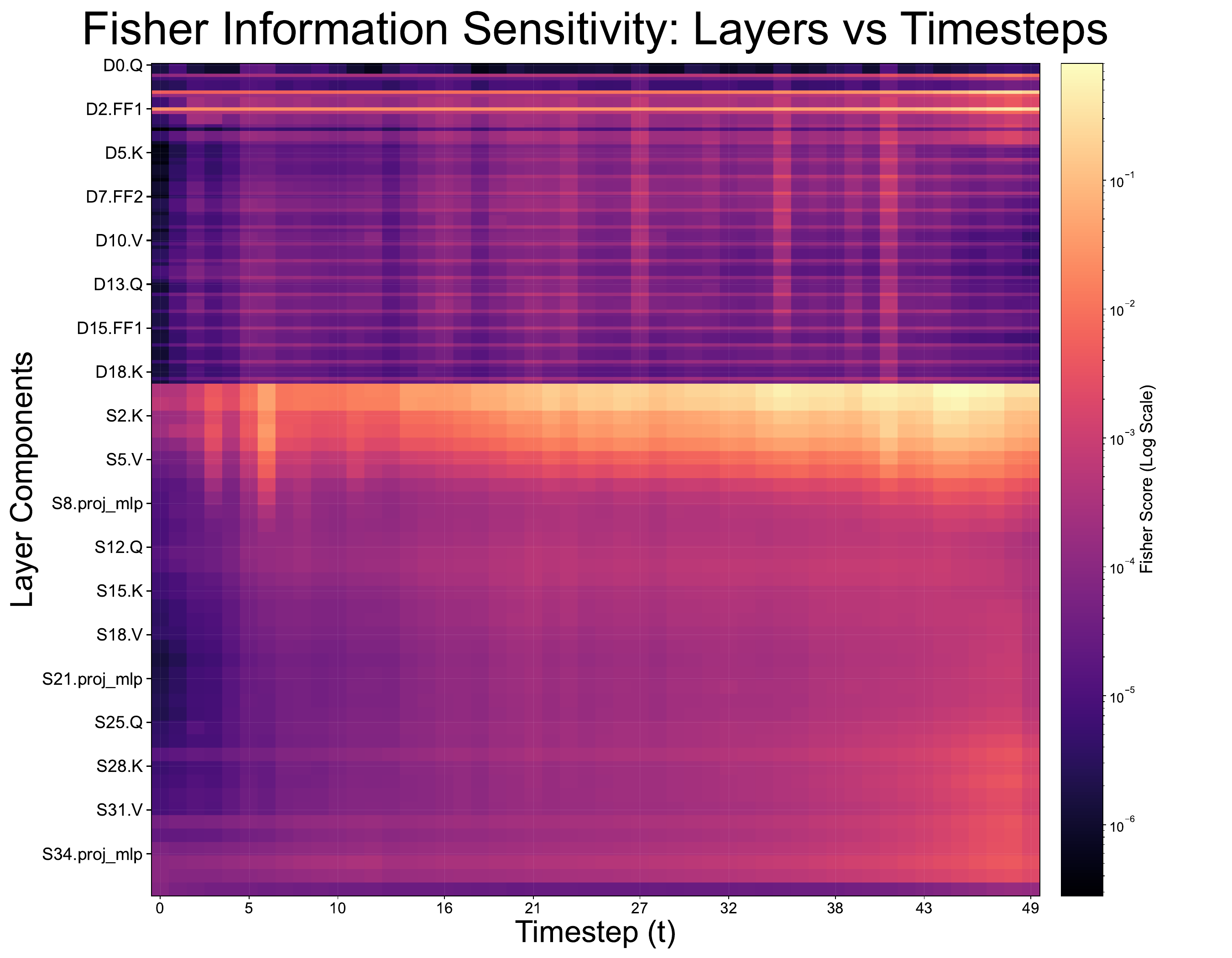} 
        % \vspace{-2mm}
        \caption{Fisher Sensitivity Heatmap}
        \label{fig:fisher_heatmap}
    \end{subfigure}
    \hfill
    \begin{subfigure}[b]{0.55\textwidth}
        \centering
        \includegraphics[width=\textwidth]{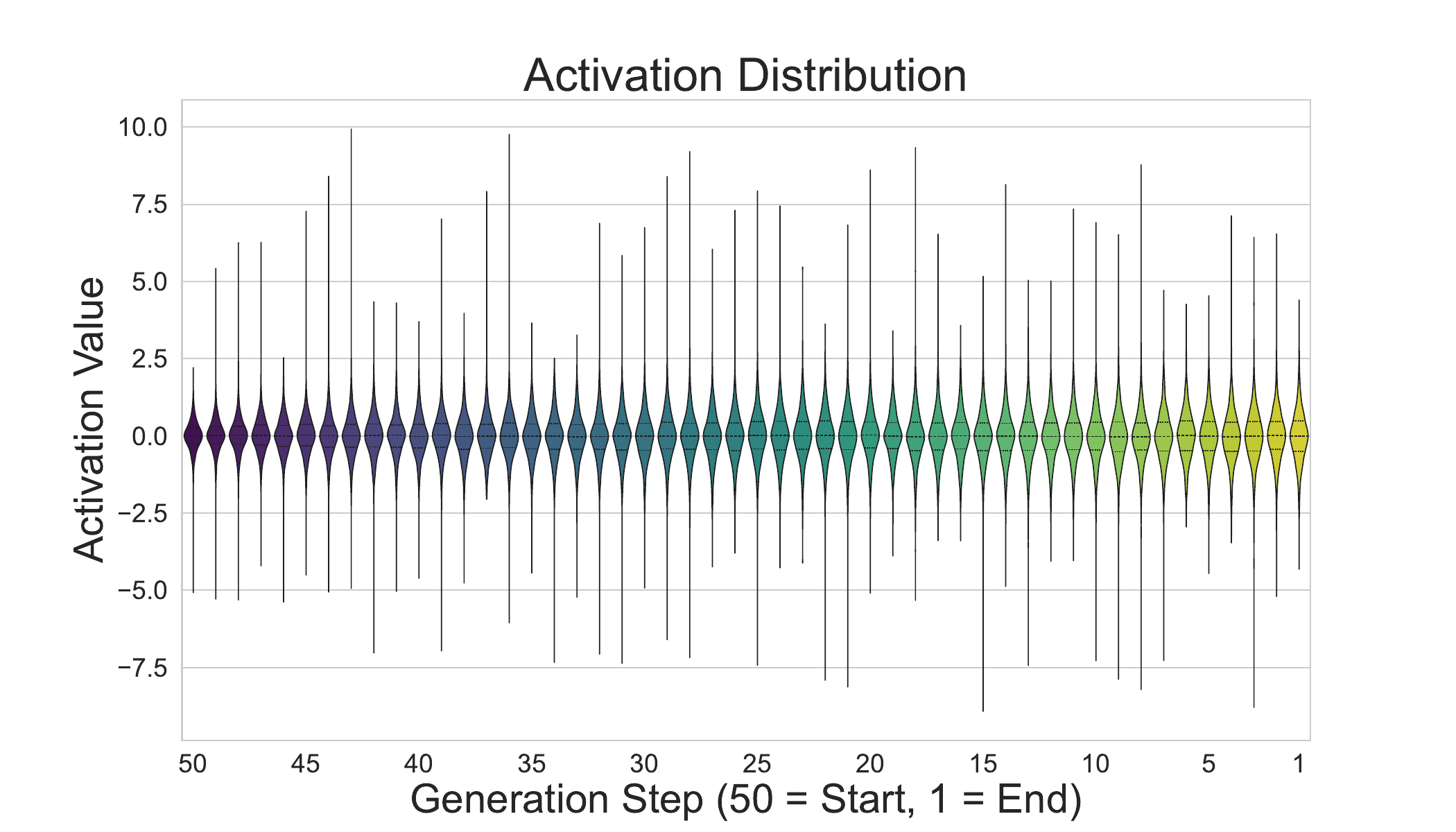} 
        % \vspace{-2mm}
        \caption{Activation Distribution Shift}
        \label{fig:activation_diff}
    \end{subfigure}
    
    \vspace{-2mm} % 调整 caption 与图的距离
    \caption{Temporal Heterogeneity in DiTs. \textbf{(a)} Normalized Fisher Information reveals that layer sensitivity varies drastically across different phases of the denoising process (e.g., structure formation vs. texture refinement). \textbf{(b)} Violin plots illustrate significant shifts in activation distributions across timesteps. These observations collectively motivate our timestep-dynamic quantization strategy.}
    \label{fig:motivation}
    \vspace{-5mm} % 调整图与正文的距离
\end{figure*}

\vspace{-3mm}
\section{Method}
\label{sec:method}
\vspace{-2mm}
\subsection{Motivation: The Temporal Heterogeneity of DiTs}
\label{sec:motivation}
\vspace{-1mm}
The core premise of PTQ methods is that layer sensitivity is relatively static or dependent solely on input content. However, in diffusion models, the input distribution evolves drastically over time, shifting from pure Gaussian noise to structured data. This raises a fundamental hypothesis: 
\vspace{-1mm}
\begin{tcolorbox}[colback=yellow!10, colframe=black, boxrule=0.5mm]
\textit{Does the sensitivity of a DiT layer to quantization noise remain constant during the denoising process?}
\end{tcolorbox}
\vspace{-1mm}
To answer this, we employ Fisher Information as a theoretically grounded metric for sensitivity. For a given layer $l$ with weights $\mathbf{W}_l$, the Fisher Information at timestep $t$ can be approximated by the expected squared gradients of the loss $\mathcal{L}$ with respect to the weights:
\begin{equation}
\label{eq:fisher_score}
\mathcal{I}_{t,l} = \mathbb{E}_{\mathbf{x}_t \sim \mathcal{D}_t} \left[ \left( \frac{\partial \mathcal{L}(\mathbf{x}_t)}{\partial \mathbf{W}_l} \right)^2 \right].
\end{equation}
Intuitively, a higher $\mathcal{I}_{t,l}$ indicates that a small perturbation (quantization noise) in layer $l$ at timestep $t$ will cause a significant increase in the final task loss.

\vspace{-1mm}
\textbf{Observation.} We visualize the normalized Fisher scores across layers and timesteps in Figure \ref{fig:fisher_heatmap}. The heatmap reveals a striking temporal heterogeneity:
(1) Layer Variance: In the same timestep, sensitivity varies significantly across different modules.
(2) Phase-Dependent Sensitivity: Certain layers are highly sensitive during the early ``structure formation'' phase but become robust to noise in the later ``texture refinement'' phase.

\vspace{-1mm}
\textbf{Implication.} This observation invalidates the optimality of uniform quantization. Assigning a static bit-width to all timesteps inevitably leads to a dilemma: it is either wasteful for insensitive steps or destructive for sensitive steps. This necessitates a Timestep-Dynamic Allocation strategy that distributes the bit-budget to where it is most needed.
\vspace{-3mm}
\subsection{Pareto-aware Timestep-Dynamic Allocation}
\label{sec:pareto_allocation}

To exploit the temporal heterogeneity revealed in Sec. \ref{sec:motivation}, we propose a pareto-aware timestep-dynamic allocation strategy that assigns specific bit-widths to each layer at each timestep. However, the search space is prohibitively large ($|\mathcal{B}|^{T \times L}$, where $|\mathcal{B}|$ is the number of candidate bit-widths). To navigate this space efficiently, we formulate the dynamic bit-width allocation problem as a constrained pathfinding task and solve it via a Pareto-aware beam search.

\subsubsection{Problem Formulation}

Let $\mathbf{b}_t = \{b_{t,l}\}_{l=1}^L$ denote the bit-width configuration vector for all layers at timestep $t$. Our goal is to find a sequence of configurations $\mathcal{P} = (\mathbf{b}_1, \dots, \mathbf{b}_T)$ that minimizes the cumulative generation error subject to an average bit-width constraint $B_{\text{target}}$. We define the step-wise reconstruction error at timestep $t$ as the MSE between the full-precision output $\epsilon_\theta(\mathbf{z}_t, t)$ and the quantized output $\hat{\epsilon}_\theta(\mathbf{z}_t, t; \mathbf{b}_t)$:
\begin{equation}
\mathcal{L}_{\text{MSE}}(\mathbf{b}_t) = \| \epsilon_\theta(\mathbf{z}_t, t) - \hat{\epsilon}_\theta(\mathbf{z}_t, t; \mathbf{b}_t) \|_2^2.
\end{equation}
The optimization problem is formulated as:
\begin{equation}
\label{eq:optimization}
\min_{\mathcal{P}} \sum_{t=1}^T \mathcal{L}_{\text{MSE}}(\mathbf{b}_t) \quad \text{s.t.} \quad \frac{1}{T \cdot L} \sum_{t=1}^T \sum_{l=1}^L b_{t,l} \le B_{\text{target}}.
\end{equation}

\vspace{-4mm}
\subsubsection{Candidate Generation}
Instead of searching bit-widths for each layer, we construct a timestep-specific set of $M$ candidate configurations $\mathcal{C}_t = \{\mathbf{c}_t^{(1)}, \dots, \mathbf{c}_t^{(M)}\}$. Crucially, these candidates are dynamically generated based on the Fisher information $\mathcal{I}_{t,l}$ at the current timestep. For example, at a sensitive timestep, the clustering algorithm may assign high precision to attention layers. Conversely, at a less sensitive step, it may assign lower precision. This adaptive approach ensures the search space aligns with the evolving sensitivity of the diffusion process, pruning irrelevant paths while enhancing efficiency.

\vspace{-2mm}
\subsubsection{Pareto-Optimal Beam Search}
\vspace{-1mm}
We employ a pareto-optimal based beam search algorithm to solve Eq. \eqref{eq:optimization}. Unlike standard beam search which keeps the top-$K$ paths based on a single metric, we maintain the Pareto Frontier of paths to balance the trade-off between cumulative error and accumulated bit-cost.

Let $\mathcal{S}_{t-1} = \{ (\text{path}_i, \text{loss}_i, \text{bits}_i) \}_{i=1}^M$ be the set of $M$ retained paths at step $t-1$. For the current step $t$, we expand each path by appending every candidate configuration $\mathbf{c} \in \mathcal{C}_t$, resulting in $M \times M$ potential new paths. For each new path $j$, we compute:
\begin{align}
\text{Cumulative Loss:} \quad & E_t^{(j)} = E_{t-1}^{(i)} + \mathcal{L}_{\text{MSE}}(\mathbf{c}), \\
\text{Cumulative Bits:} \quad & B_t^{(j)} = B_{t-1}^{(i)} + \text{avg}(\mathbf{c}).
\end{align}
From these $M^2$ candidates, we select the $M$ paths that form the Pareto Frontier in the (Bits, Loss) plane. A path $j$ dominates path $k$ if $B_t^{(j)} \le B_t^{(k)}$ and $E_t^{(j)} \le E_t^{(k)}$. We retain only the non-dominated paths (or the top-$M$ closest to the origin if the frontier size $> M$). This process iterates from $t=1$ to $T$, yielding a final set of paths from which we select the one satisfying $B_{\text{target}}$.
\vspace{-1mm}
\subsubsection{Final Selection}
\vspace{-1mm}
Upon completing the beam search at $t=T$, we obtain a set of $M$ Pareto-optimal paths. While the search relies on the cumulative MSE as a computationally efficient proxy, MSE does not always perfectly align with human perception or generative metrics. To ensure the final configuration delivers superior perceptual quality, we perform a lightweight end-to-end generation test on the surviving $M$ paths. We generate a small batch of samples using each candidate path and evaluate them against task-specific metrics (\textit{e.g.}, CLIP score for text-to-image alignment). The path yielding the best trade-off between the generative metric and the bit-budget is selected as the final quantization policy. This two-stage strategy—efficient MSE-based pruning followed by accurate metric-based selection—drastically reduces the search overhead while guaranteeing optimal generation quality.

\begin{figure}[t!]
    \centering
    \includegraphics[width=\columnwidth]{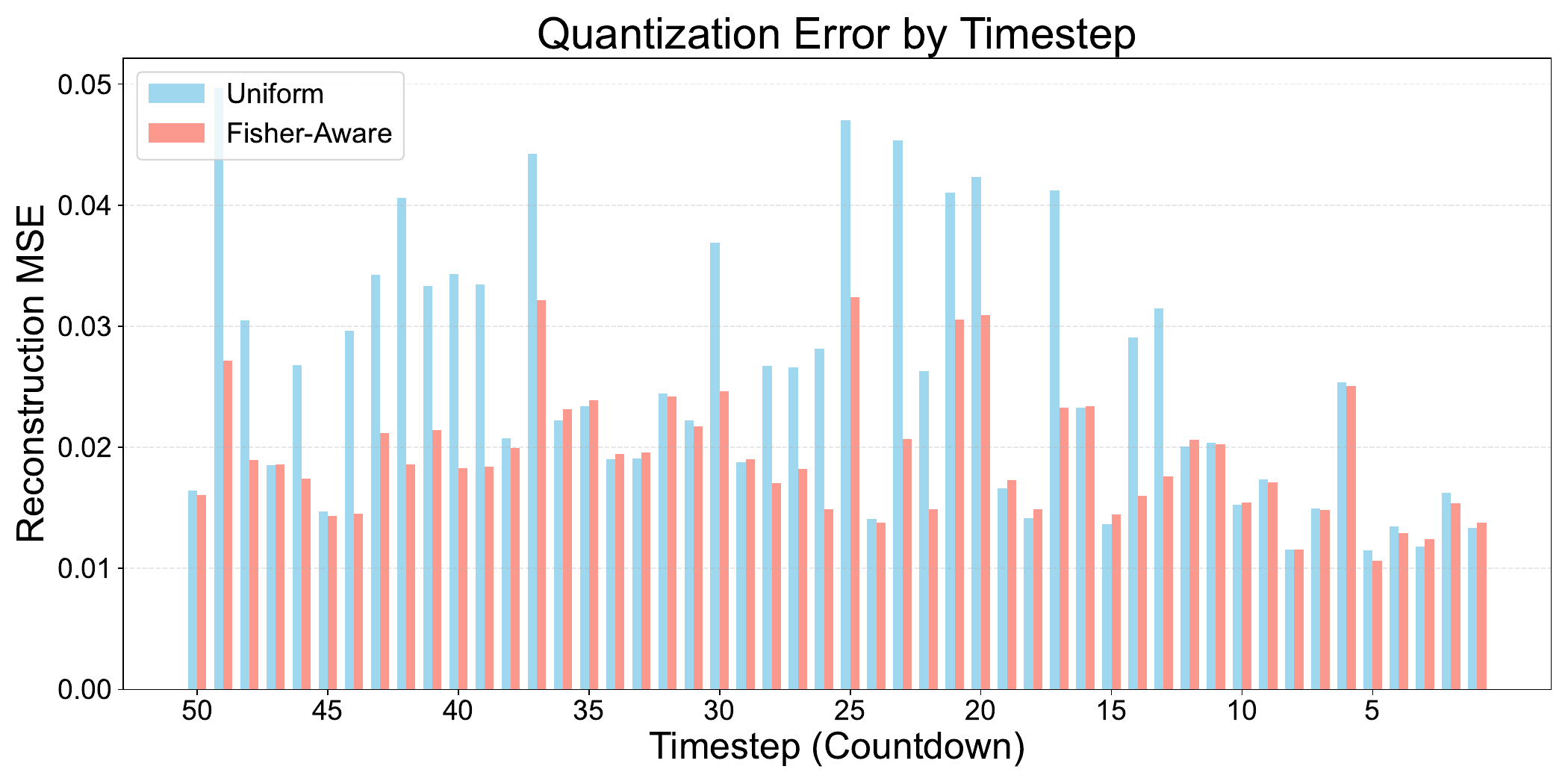}
    \vspace{-5mm}
    \caption{Reconstruction error of transformer.blocks.18.ff.net.2 on Flux-Dev across timesteps. Compared to standard Uniform Calibration (Blue), our Fisher-Guided Calibration (Pink) effectively suppresses quantization noise in high-sensitivity regions, resulting in a smoother and lower risk profile.}
    \vspace{-5mm}
    \label{fig:weight_error_comparison}
    \vspace{-6mm}
\end{figure}

\begin{table*}[ht!]
    \centering
    \scriptsize
    \caption{Quantization results on GenEval. The best and the second best results are marked with {\color{red}red} and {\color{blue}blue}, respectively.}
    \label{tab:geneval_results}
    \vspace{-2mm}
    \setlength{\tabcolsep}{2.0mm}
    \resizebox{\textwidth}{!}{%
    \begin{tabular}{c|c|c|ccccccc}
    
\hline
\toprule[0.15em]
\rowcolor{colorhead}
Model & Method & Bits (W/A) & Single Object & Position & Counting & Two Object & Colors & Color Attribute & Total \\
\midrule[0.15em]

% ========== FLUX-dev 部分 (6行) ==========
\multirow{9}{*}{\makecell{FLUX-dev\\(50 steps)}} 
& FP              & 16/16 & 0.9844 & 0.2025 & 0.7688 &	0.8232 & 0.7713 & 0.4500 & 0.6667  \\
\cmidrule{2-10}
& Q-DiT           &  4/4  & 0.0469 & 0.0000 & 0.0094 & 0.0000 & 0.0239 & 0.0025 & 0.0138 \\
& SmoothQuant     &  4/4  & 0.0025 & 0.0000 & 0.0031 & 0.0000 & 0.0106 & 0.0000 & 0.0065 \\
& Quarot          &  4/4  & 0.6344 & 0.0250 & 0.2938 & 0.1061 & 0.3457 & 0.0500 & 0.2425 \\
& ViDiT-Q         &  4/4  & 0.7094 & 0.0275 & 0.3250 & 0.1465 & 0.4096 & 0.0600 & 0.2797 \\
& SVDQuant        &  4/4  & {\color{red}0.9812} & {\color{blue}0.1400} & 0.6100 & {\color{blue}0.7100} & {\color{blue}0.6979} & {\color{blue}0.3000} & {\color{blue}0.5732} \\
% \cellcolor{colorours}
& AdaTSQ (ours)   &  4/4  & {\color{red}0.9812} & {\color{red}0.2600} & {\color{red}0.6400} & {\color{red}0.7700} & {\color{red}0.7083} & {\color{red}0.3500} & {\color{red}0.6183} \\
& SVDQuant        &  3/3  & 0.8688 & 0.0325 & 0.4250 & 0.2879 & 0.5239 & 0.1225 & 0.3768 \\
& AdaTSQ (ours)   &  3/3  & {\color{blue}0.9562} & 0.0825 & {\color{blue}0.6281} & 0.5480 & 0.6559 & 0.2900 & 0.5270 \\
\midrule[0.15em]

% ========== FLUX-schnell 部分 (6行) ==========
\multirow{9}{*}{\makecell{FLUX-schnell\\(4 steps)}} 
& FP              & 16/16 & 0.9969 & 0.2750 & 0.6000 & 0.8838 & 0.7420 & 0.4875 & 0.6642 \\
\cmidrule{2-10}
& Q-DiT           &  4/4  & 0.7406 & 0.0725 & 0.3781 & 0.4242 & 0.4176 & 0.2075 & 0.3734 \\ 
& SmoothQuant     &  4/4  & 0.6188 & 0.0425 & 0.2719 & 0.2929 & 0.3165 & 0.1425 & 0.2808 \\
& Quarot          &  4/4  & 0.8188 & 0.1175 & 0.4719 & 0.5429 & 0.5186 & 0.2750 & 0.4575 \\
& ViDiT-Q         &  4/4  & 0.8875 & 0.1300 & 0.5156 & 0.5859 & 0.5851 & 0.2675 & 0.4953 \\
& SVDQuant        &  4/4  & {\color{blue}0.9938} & {\color{blue}0.2600} & 0.4500 & {\color{blue}0.9100} & {\color{blue}0.7083} & 0.4200 & 0.6237 \\
% \cellcolor{colorours}
& AdaTSQ (ours)   &  4/4  & {\color{red}0.9969} & {\color{red}0.2775} & {\color{red}0.6219} & 0.8939 & {\color{red}0.7926} & {\color{red}0.4975} & {\color{red}0.6801} \\

& SVDQuant        &  3/3  & 0.8200 & 0.1600 & 0.4656 & 0.6470 & 0.5601 & 0.3725 & 0.5042 \\
% \cellcolor{colorours}
& AdaTSQ (ours)   &  3/3  & {\color{red}0.9969} & 0.2300 & {\color{blue}0.5300} & {\color{red}0.9200} & 0.6875 & {\color{blue}0.4400} & {\color{blue}0.6341} \\
\midrule[0.15em]

% ========== Z-Image 部分 (6行) ==========
\multirow{9}{*}{\makecell{Z-Image\\(10 steps)}} 
& FP              & 16/16 & 1.0000 & 0.4675 & 0.7094 & 0.9066 & 0.8590 & 0.5825 & 0.7542 \\
\cmidrule{2-10}
& Q-DiT           &  4/4  & N/A & N/A & N/A & N/A & N/A & N/A & N/A \\
& SmoothQuant     &  4/4  & 0.0031 & 0.0000 & 0.0000 & 0.0000 & 0.0000 & 0.0000 & 0.0000 \\
& Quarot          &  4/4  & 0.9062 & 0.2500 & 0.4156 & 0.5051 & 0.6915 & 0.3425 & 0.5185 \\
& ViDiT-Q         &  4/4  & {\color{blue}0.9719} & 0.4350 & 0.5844 & 0.7045 & 0.7766 & {\color{blue}0.5325} & 0.6675 \\
& SVDQuant        &  4/4  & {\color{red}0.9938} & {\color{blue}0.4850} & {\color{blue}0.6325} & 0.8430 & 0.8332 & 0.5200 & {\color{blue}0.7179} \\
& AdaTSQ (ours)   &  4/4  & {\color{red}0.9938} & {\color{red}0.5200} & {\color{red}0.6812} & {\color{red}0.8914} & {\color{blue}0.8723} & {\color{red}0.6125} & {\color{red}0.7619} \\
& SVDQuant        &  3/3  & 0.0047 & 0.0000 & 0.0000 & 0.0000 & 0.0000 & 0.0000 & 0.0001 \\
& AdaTSQ (ours)   &  3/3  & {\color{red}0.9938} & 0.4100 & 0.5500 & {\color{blue}0.8700} & {\color{red}0.8854} & 0.4545 & 0.6940 \\
\bottomrule[0.15em]

    \end{tabular}
    }
\vspace{-2mm}
\end{table*}
\begin{table*}[t!]
    \centering
    \scriptsize
    \caption{Quantization results on VBench. The best and the second best results are marked with {\color{red}red} and {\color{blue}blue}, respectively. }
    \label{tab:vbench_results}
    \vspace{-2mm}
    \setlength{\tabcolsep}{2mm}
    \resizebox{\textwidth}{!}{%
    \begin{tabular}{c|c|c|cccccccc}
    
\hline
\toprule[0.15em]
\rowcolor{colorhead}
& & Bits  & Imaging & Aesthetic & Motion & Dynamic & BG.      & Scene    & Overall & Subject \\
\rowcolor{colorhead}
\multirow{-2}{*}{Model} & \multirow{-2}{*}{Methods} & (W/A) & Quality & Quality   & Smooth.& Degree  & Consist. & Consist. & Consist. & Consist. \\
\midrule[0.15em]
\multirow{7}{*}{\makecell{Wan2.1-1.3B\\(25 steps)}} 
& FP              & 16/16 & 0.6055 & 0.5643 & 0.9558 & 0.8333 & 0.9547 & 0.2863 & 0.2510 & 0.9222 \\
\cmidrule{2-11}
& Q-DiT           &  4/4  & 0.4302 & 0.3070 & 0.9339 &    N/A   & 0.9515 &    N/A   & 0.0267 & {\color{blue}0.9062} \\
& SmoothQuant     &  4/4  & {\color{blue}0.5266} & {\color{blue}0.3838} & 0.8517 &    N/A   & 0.9519 &    N/A   & 0.0120 & 0.9000 \\
& Quarot          &  4/4  & 0.3840 & 0.3415 & 0.9469 & 0.5833 & {\color{blue}0.9631} & 0.0094 & {\color{blue}0.1283} & 0.8986 \\
& ViDiT-Q         &  4/4  & 0.4893 & 0.2487 & 0.9445 &    N/A   & 0.9500 &    N/A   & 0.0254 & 0.8760 \\
& SVDQuant        &  4/4  & 0.4073 & 0.3169 & {\color{blue}0.9485} & {\color{blue}0.6444} & 0.9562 & {\color{blue}0.0203} & 0.1024 & 0.9041 \\
% \cellcolor{colorours}
& AdaTSQ (ours)   &  4/4  & {\color{red}0.5626} & {\color{red}0.5073} & {\color{red}0.9510} & {\color{red}0.8472} & {\color{red}0.9658} & {\color{red}0.2253} & {\color{red}0.2356} & {\color{red}0.9078} \\
\bottomrule[0.15em]

    \end{tabular}
    } % resize box
    \vspace{-5mm}
\end{table*}

\vspace{-1mm}
\subsection{Fisher-Guided Temporal Calibration}
\label{sec:fisher_calibration}
\vspace{-1mm}
While Sec. \ref{sec:pareto_allocation} resolves activation quantization via timestep-wise bit-width dynamic allocation, weights must remain static across timesteps to avoid memory overhead. Finding a single quantized configuration $\hat{\mathbf{W}}$ that is robust across all temporal phases is challenging. Existing methods like ViDiT-Q \cite{zhao2024viditq} (no calibration) or SVDQuant \cite{li2024svdquant} (uniform calibration) overlook varying temporal sensitivity, often leading to structural artifacts. To address this, we propose \textbf{Fisher-Guided Temporal Calibration} (Figure \ref{fig:overview}). This approach modifies the standard Hessian-based objective to explicitly prioritize critical timesteps based on their Fisher information.

\vspace{-3mm}
\subsubsection{Temporal Importance Modeling}
\vspace{-1mm}
To rigorously ground our strategy, we view weight quantization as an Expected Risk Minimization problem. Let $\mathcal{L}(\theta)$ be the task loss. The loss perturbation due to quantization error $\Delta \mathbf{W}_l$ can be approximated via Taylor expansion: $\mathbb{E}[\Delta \mathcal{L}] \approx \frac{1}{2} \Delta \mathbf{W}_l^\top \mathbf{H}_{\mathcal{L}, l} \Delta \mathbf{W}_l$. Standard PTQ approximates the Hessian $\mathbf{H}_{\mathcal{L}, l}$ using input covariance $\mathbf{X}\mathbf{X}^\top$, implicitly assuming uniform risk across time.

However, our analysis in Sec. \ref{sec:motivation} reveals that the curvature of the loss landscape varies drastically across time and layers. To align the calibration objective with the true risk, we introduce a layer-wise temporal re-weighting mechanism. We define the temporal importance $\alpha_{t,l}$ using Fisher Information $\mathcal{I}_{t,l}$ as a proxy for local curvature magnitude.

To prevent distribution collapse—where a few dominant timesteps suppress others—we employ a Temperature-Scaled Softmax normalization for each layer:
\begin{equation}
\alpha_{t,l} = \frac{\exp(\mathcal{I}_{t,l} / \tau)}{\sum_{t'=1}^T \exp(\mathcal{I}_{t',l} / \tau)},
\end{equation}
where $\tau$ is a temperature hyperparameter. This formulation ensures that each layer focuses on its own most critical timesteps during calibration.

\vspace{-1mm}
\subsubsection{Risk-Aware Hessian Calibration}
\vspace{-1mm}
Incorporating $\alpha_{t,l}$, we reformulate the quantization objective to minimize the Temporally Weighted Risk:
\begin{equation}
\label{eq:weighted_obj}
\min_{\hat{\mathbf{W}}_l} \sum_{t=1}^T \alpha_{t,l} \|\mathbf{W}_l\mathbf{X}_{t,l} - \hat{\mathbf{W}}_l\mathbf{X}_{t,l}\|_2^2.
\end{equation}
Here, $\mathbf{X}_{t,l}$ denotes calibration inputs at timestep $t$. To solve this efficiently, we derive the Risk-Aware Hessian $\mathbf{H}'_l$:
\begin{equation}
\mathbf{H}'_l = \sum_{t=1}^T \alpha_{t,l} (\mathbf{X}_{t,l} \mathbf{X}_{t,l}^\top).
\end{equation}
This is equivalent to scaling input features $\mathbf{X}_{t,l}$ by $\sqrt{\alpha_{t,l}}$ prior to Hessian accumulation. By doing so, we seamlessly inject temporal awareness into standard solvers, effectively warping the optimization landscape to preserve structural integrity where it matters most for each component.

\begin{figure*}[t]
    \centering
    \setlength{\tabcolsep}{1pt} 
    \renewcommand{\arraystretch}{0.5} 

    \newcommand{\imgwidth}{0.108\textwidth} 

    \begin{tabular}{ccccccccc}
        % --- Header Row (Method Names) ---
        \tiny FP16 & \tiny Q-DiT (W4A4) & \tiny SmoothQ (W4A4) & \tiny Quarot (W4A4) & \tiny ViDiT-Q (W4A4) & \tiny SVDQ (W4A4) & \tiny AdaTSQ (W4A4) & \tiny SVDQ (W3A3) & \tiny AdaTSQ (W3A3) \\
        
        % --- Row 1 Images ---
        \includegraphics[width=\imgwidth]{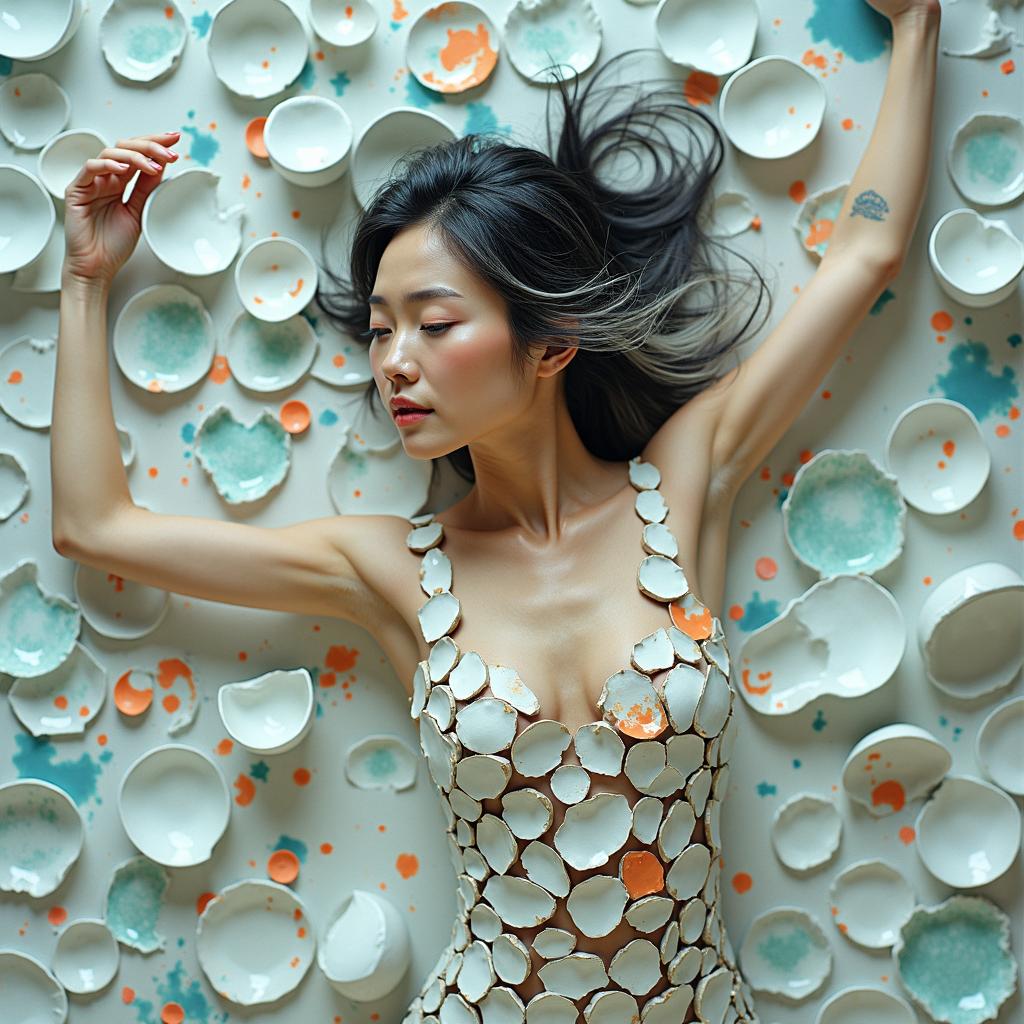} &
        \includegraphics[width=\imgwidth]{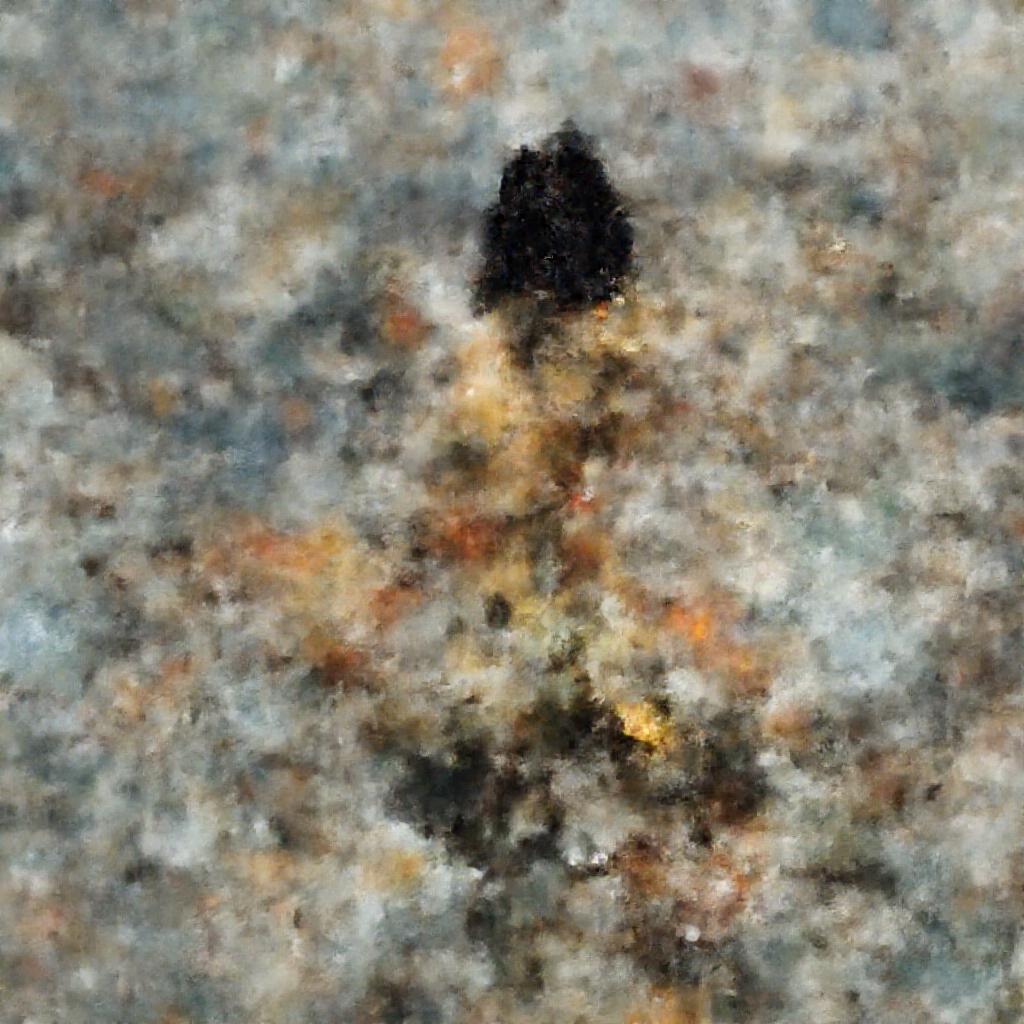} &
        \includegraphics[width=\imgwidth]{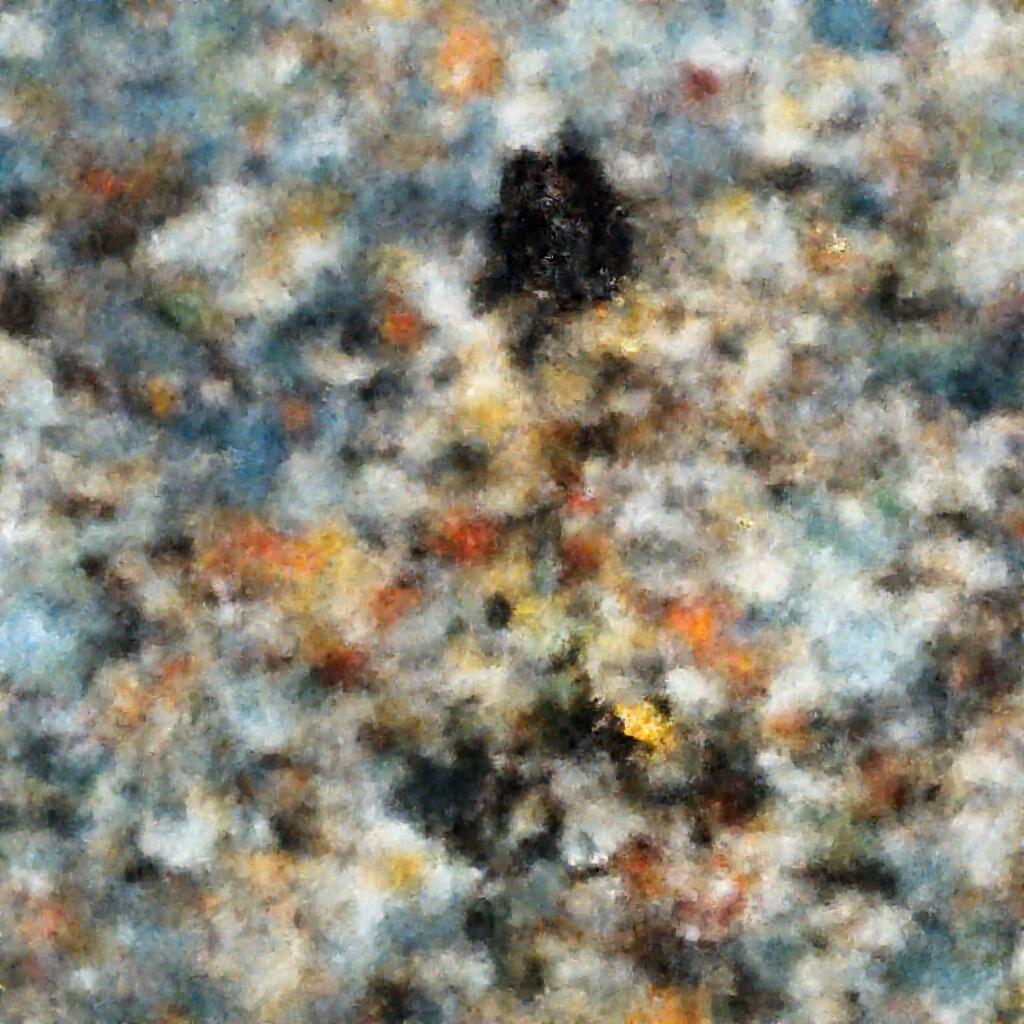} &
        \includegraphics[width=\imgwidth]{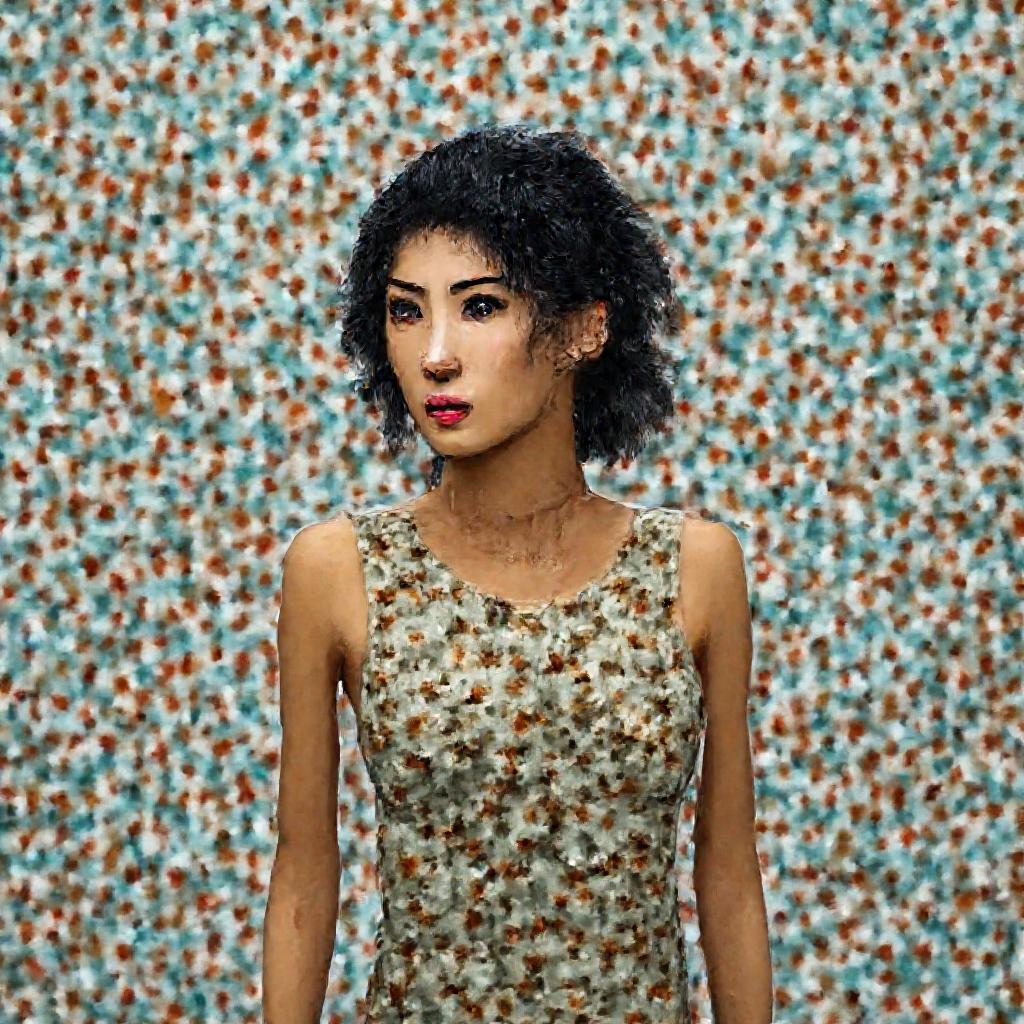} &
        \includegraphics[width=\imgwidth]{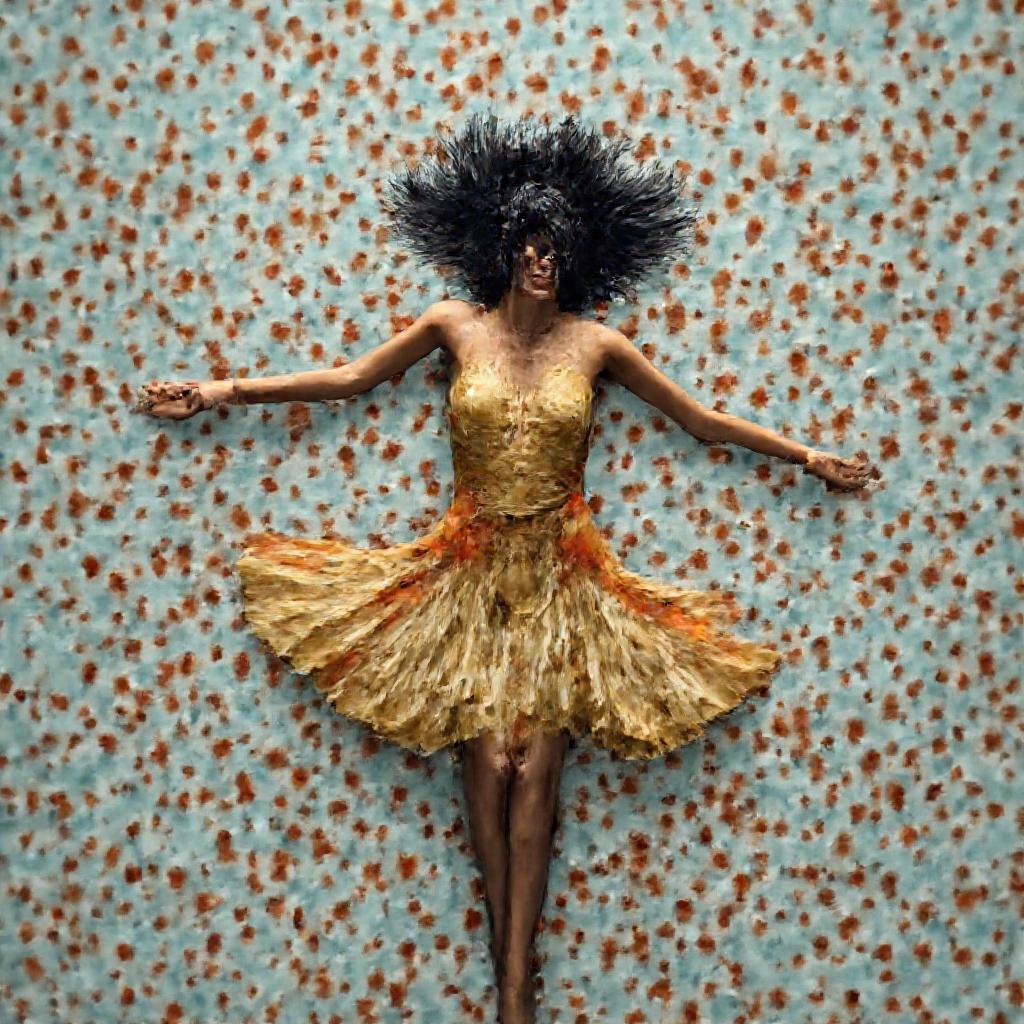} &
        \includegraphics[width=\imgwidth]{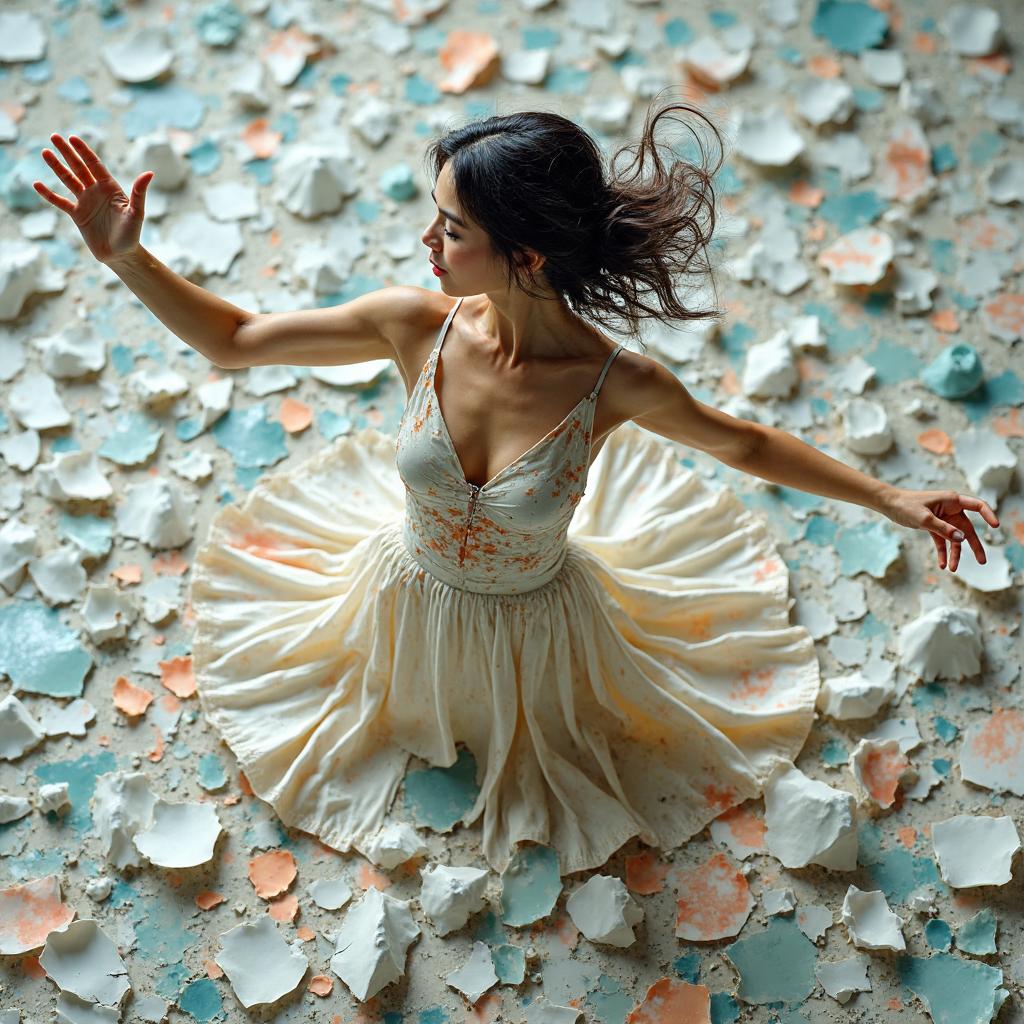} &
        \includegraphics[width=\imgwidth]{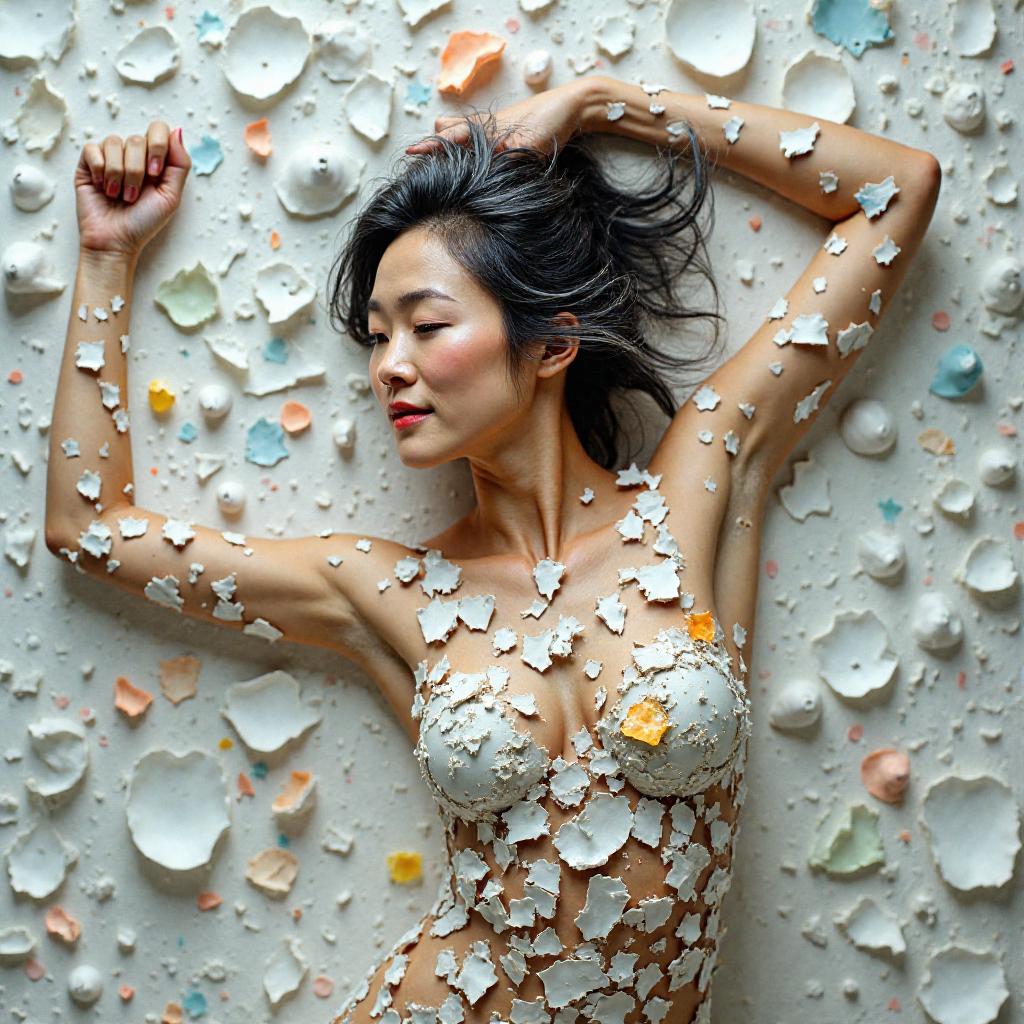} &
        \includegraphics[width=\imgwidth]{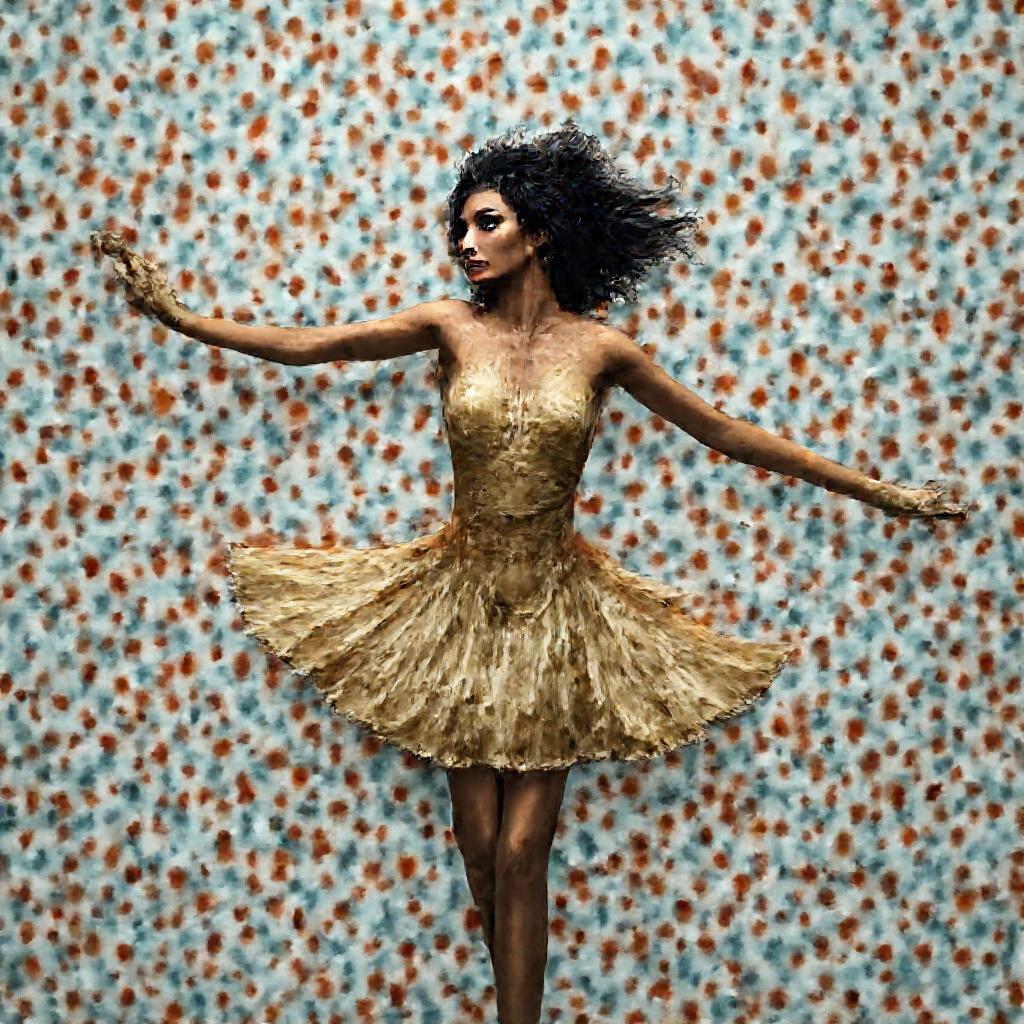} &
        \includegraphics[width=\imgwidth]{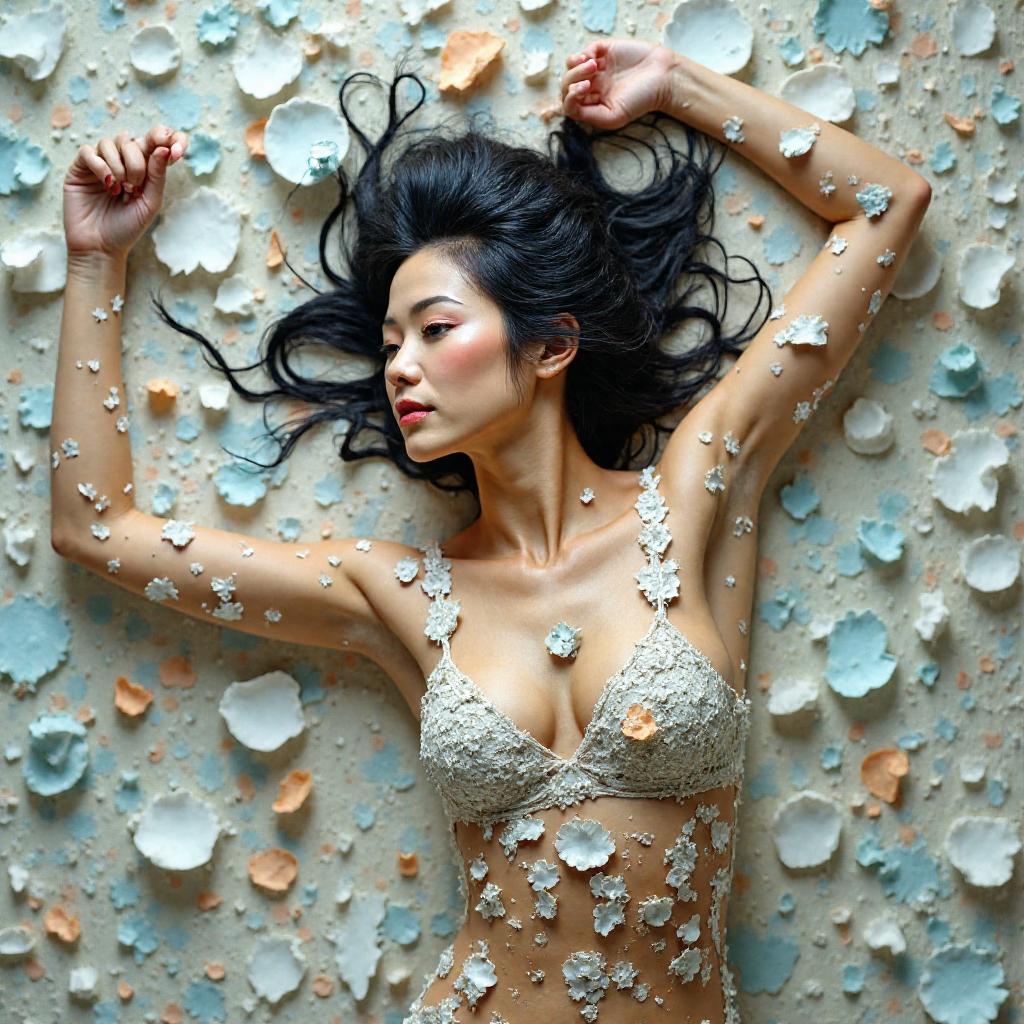} \\
        
        % --- Row 1 Prompt (跨9列) ---
        \multicolumn{9}{c}{\parbox{0.95\textwidth}{\scriptsize Model: Flux-Dev. \textit{Prompt: A middle-aged woman of Asian descent, her dark hair streaked with silver, appears fractured and splintered...}}} \\
        \noalign{\vskip 1mm} % 增加一点间距

        % --- Row 2 Images ---
        \includegraphics[width=\imgwidth]{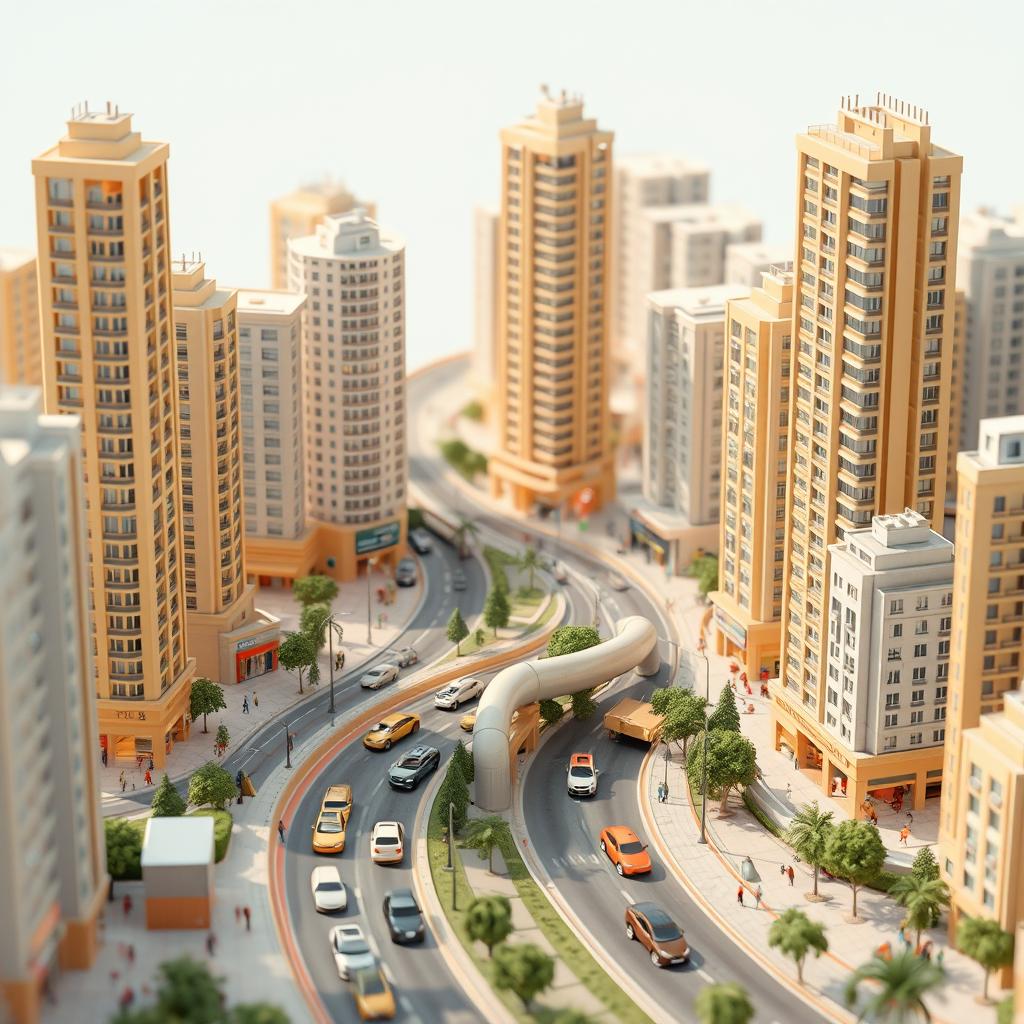} &
        \includegraphics[width=\imgwidth]{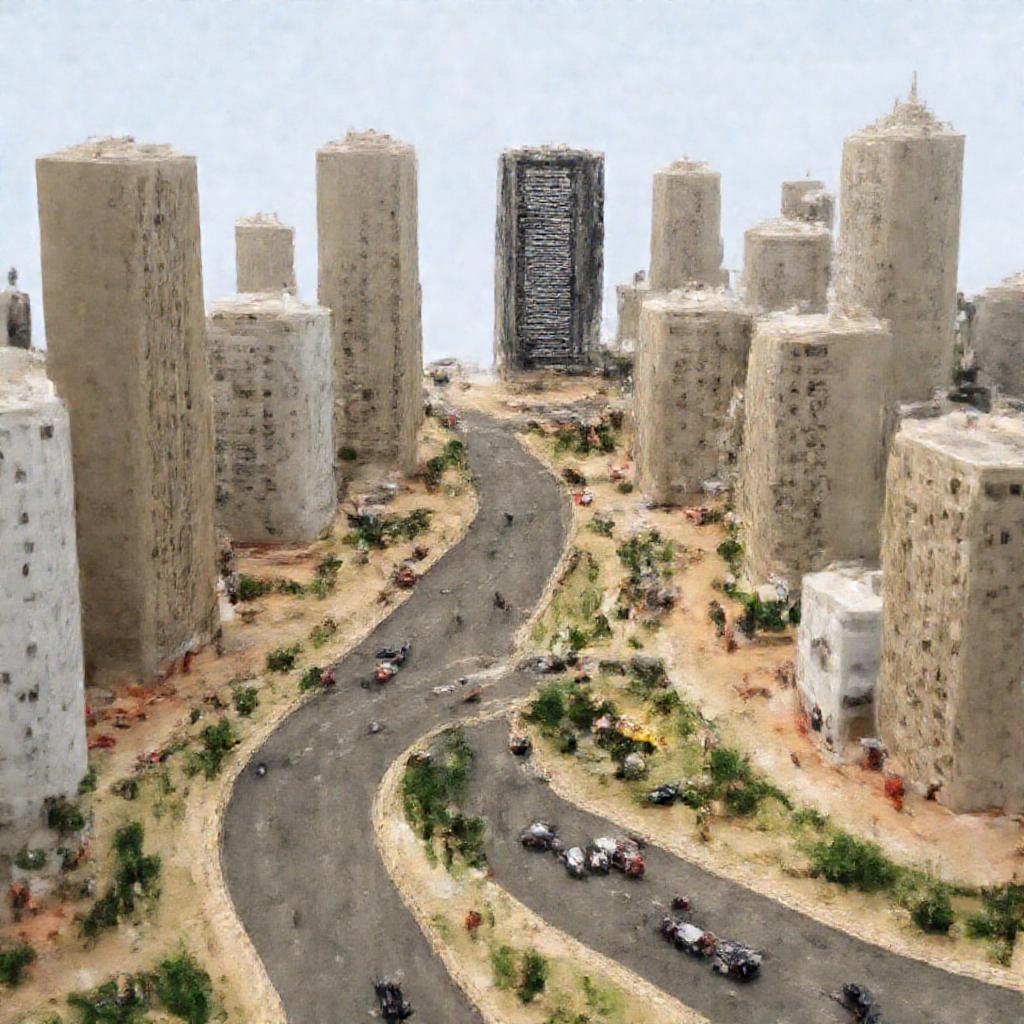} &
        \includegraphics[width=\imgwidth]{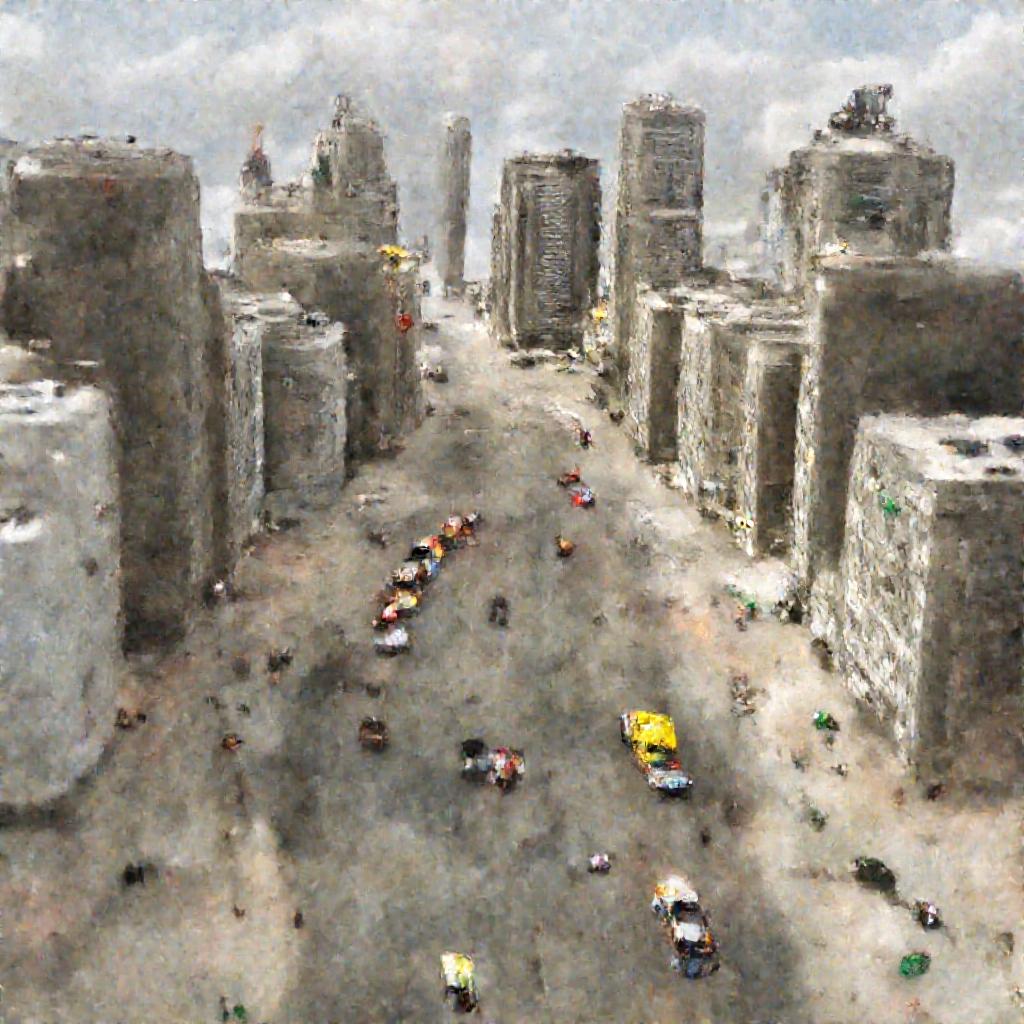} &
        \includegraphics[width=\imgwidth]{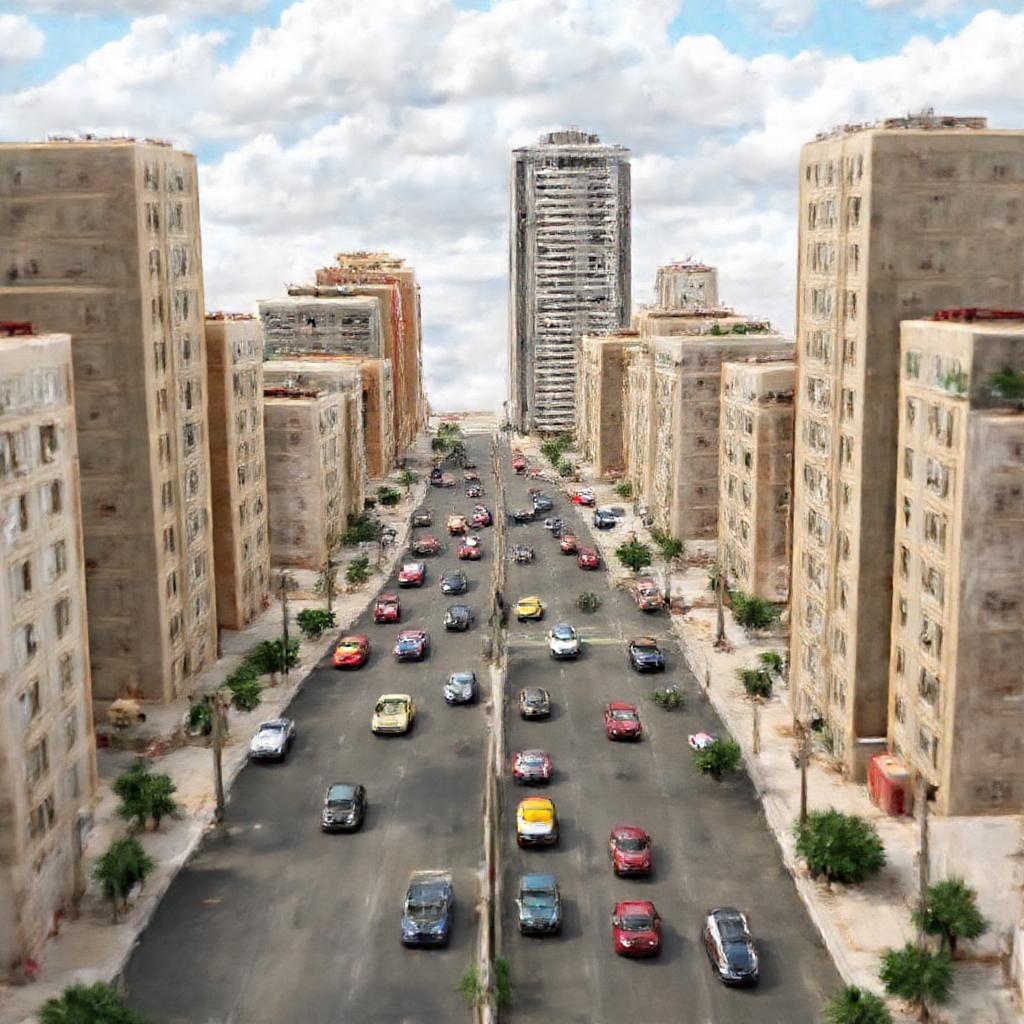} &
        \includegraphics[width=\imgwidth]{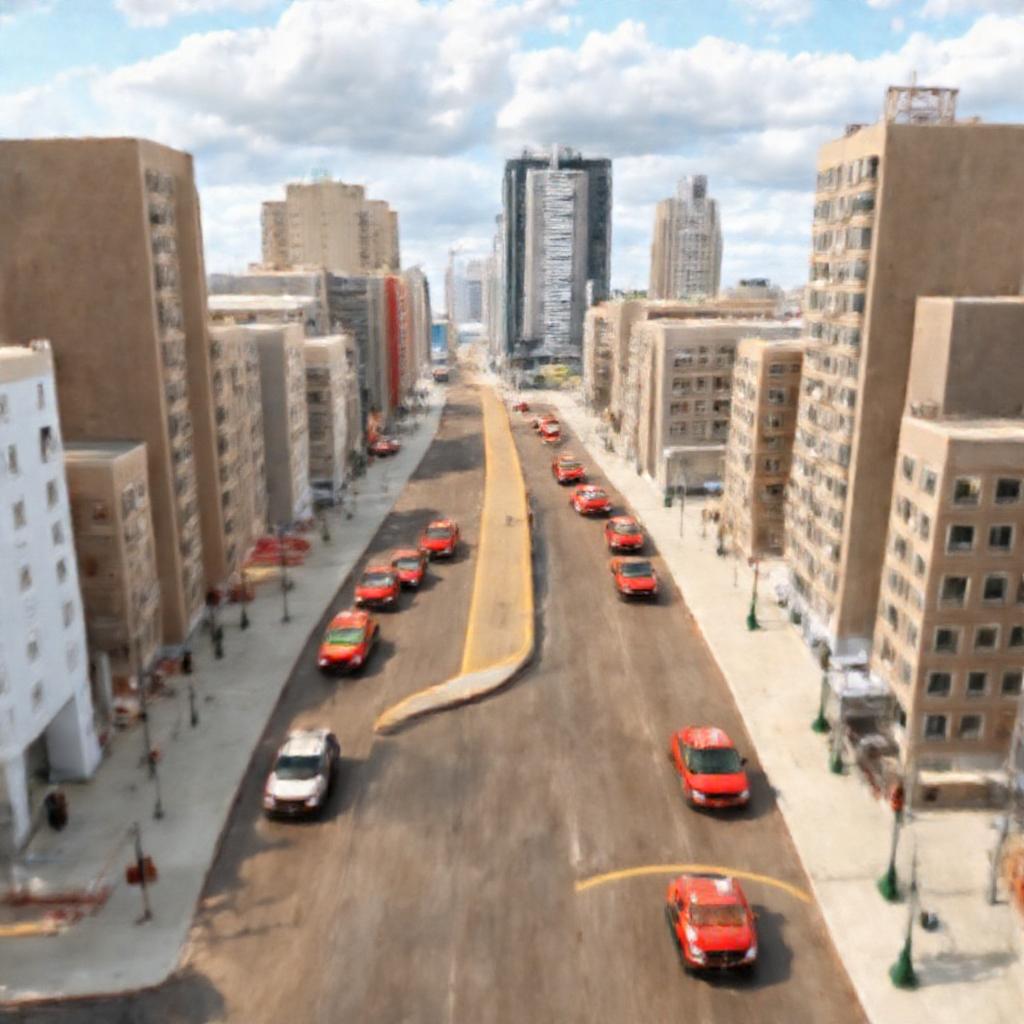} &
        \includegraphics[width=\imgwidth]{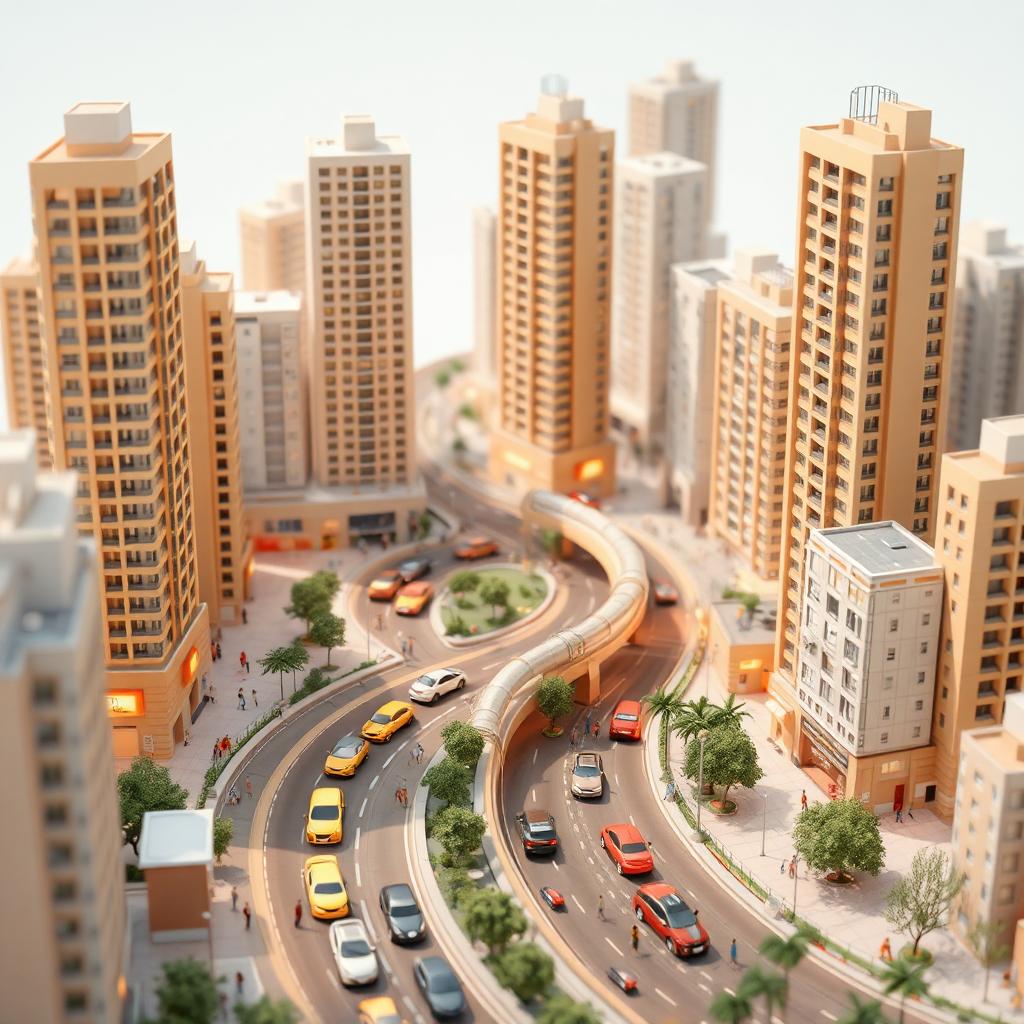} &
        \includegraphics[width=\imgwidth]{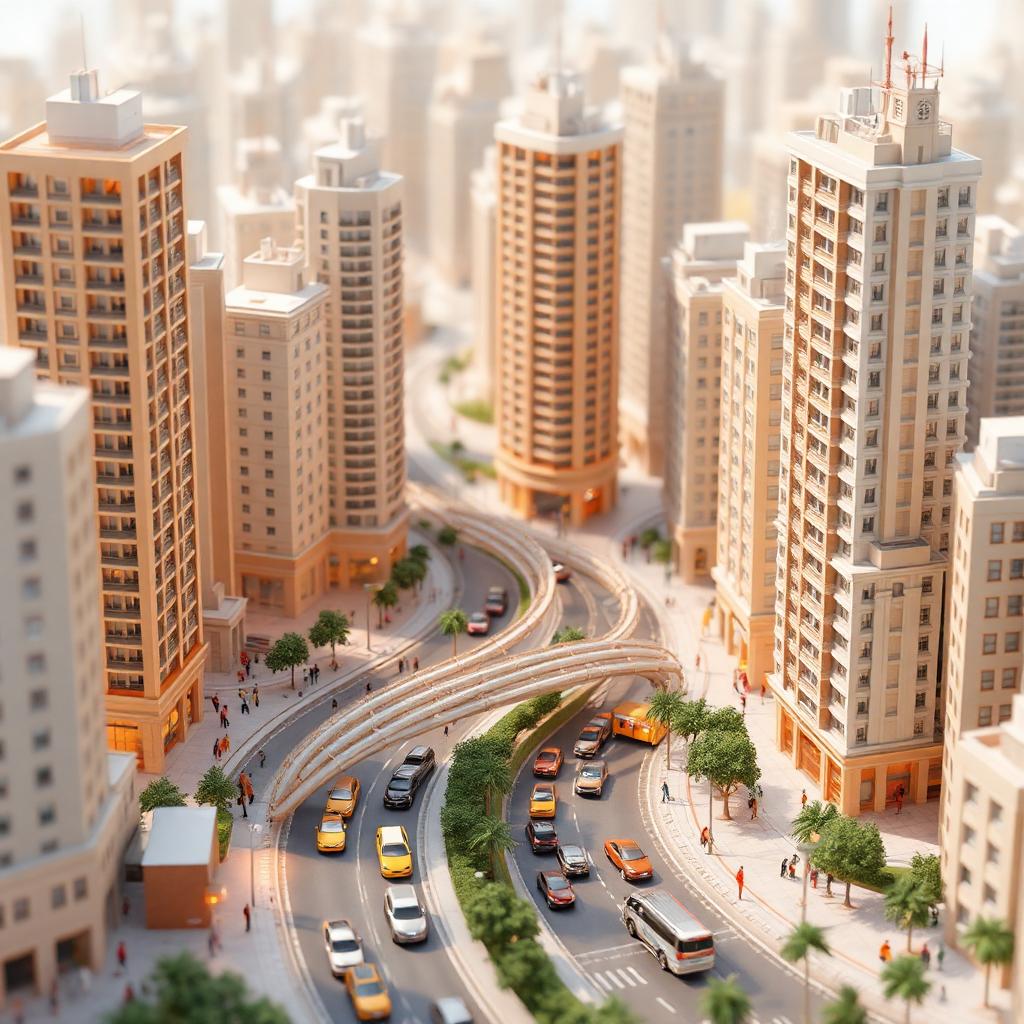} &
        \includegraphics[width=\imgwidth]{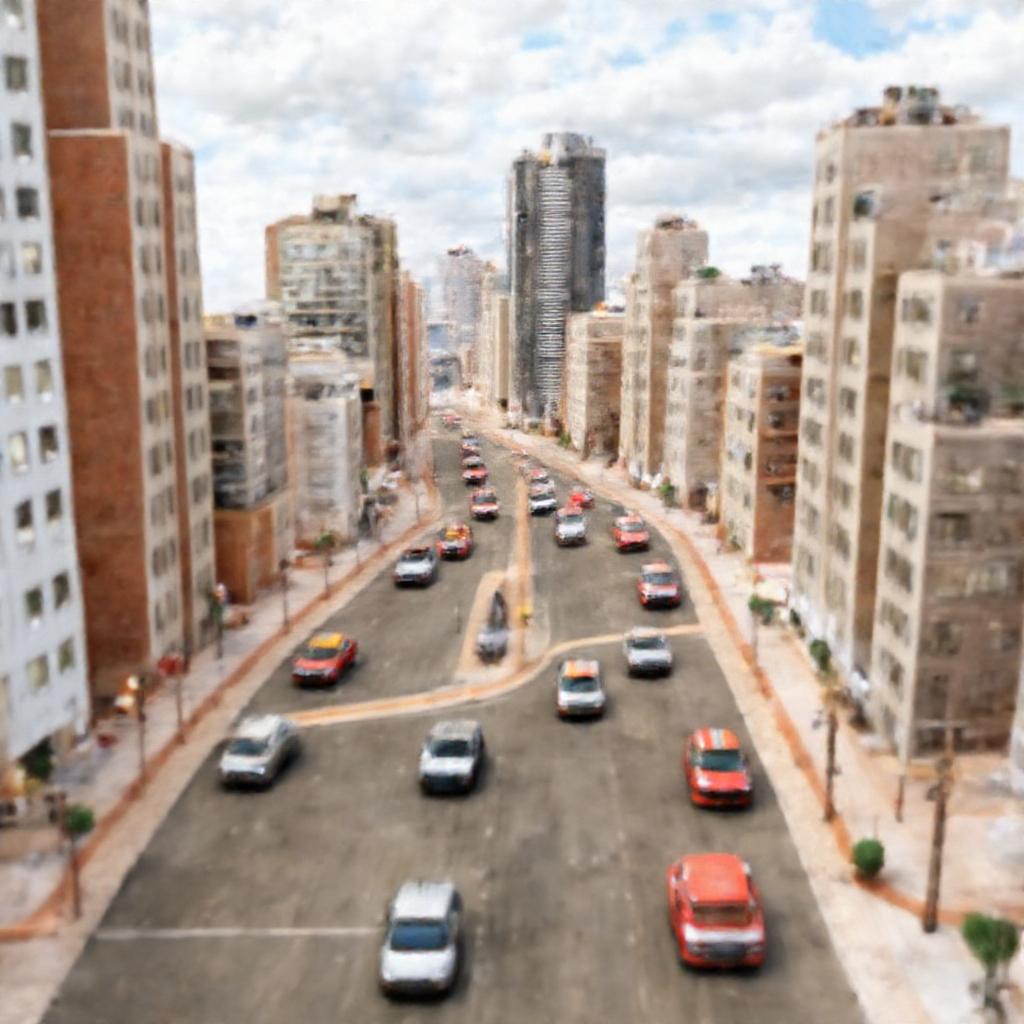} &
        \includegraphics[width=\imgwidth]{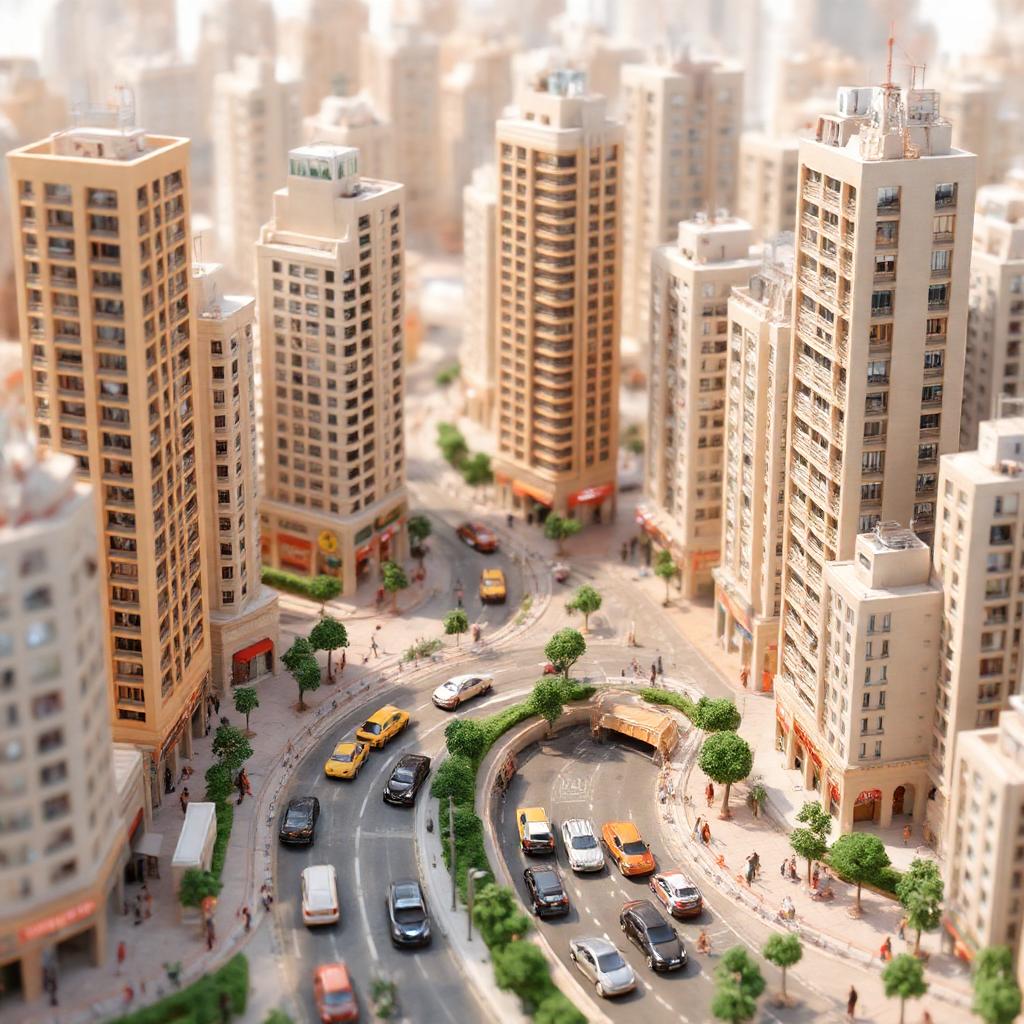} \\

        % --- Row 2 Prompt ---
        \multicolumn{9}{c}{\parbox{0.95\textwidth}{\scriptsize Model: Flux-Schnell.    \textit{Prompt: 3D rendering miniature scene design, Many tall buildings, A winding urban road runs through the middle...}}} \\
        \noalign{\vskip 1mm}

        % --- Row 3 Images ---
        \includegraphics[width=\imgwidth]{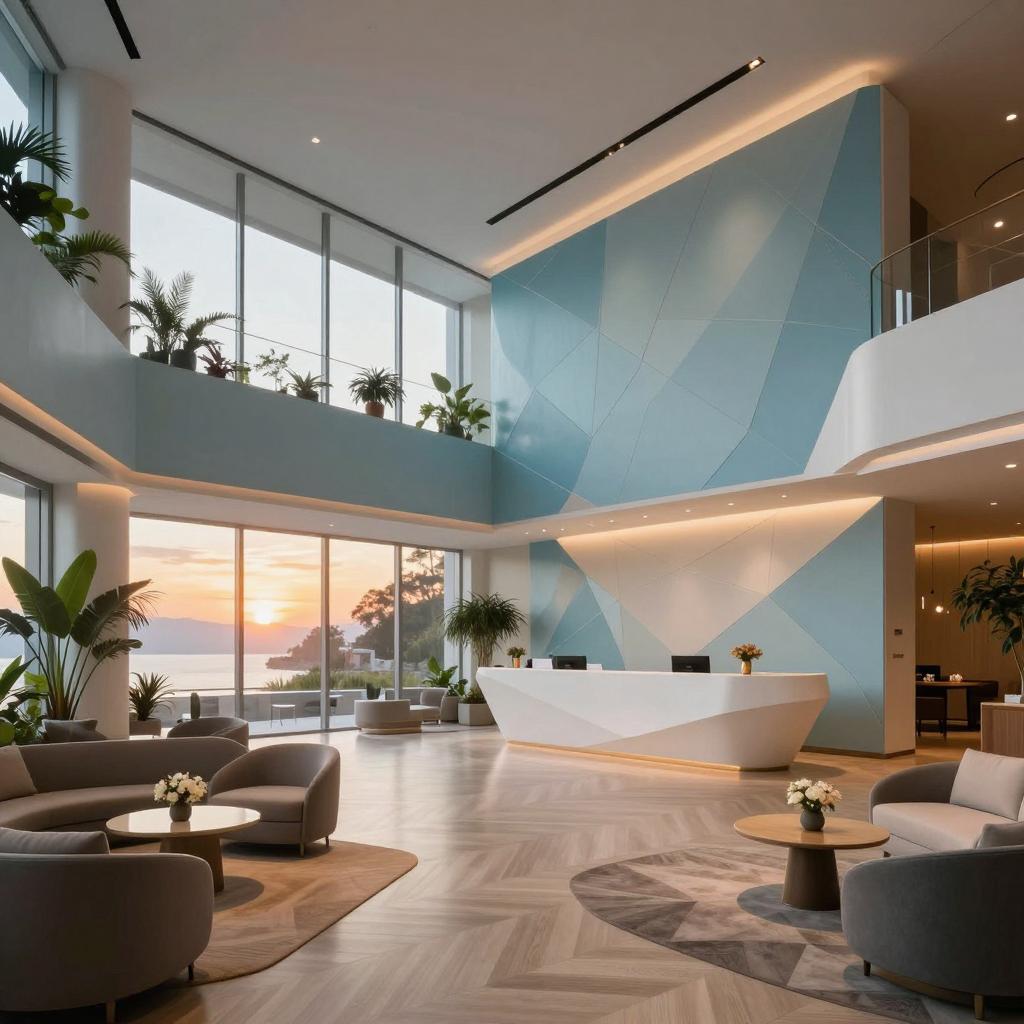} &
        \includegraphics[width=\imgwidth]{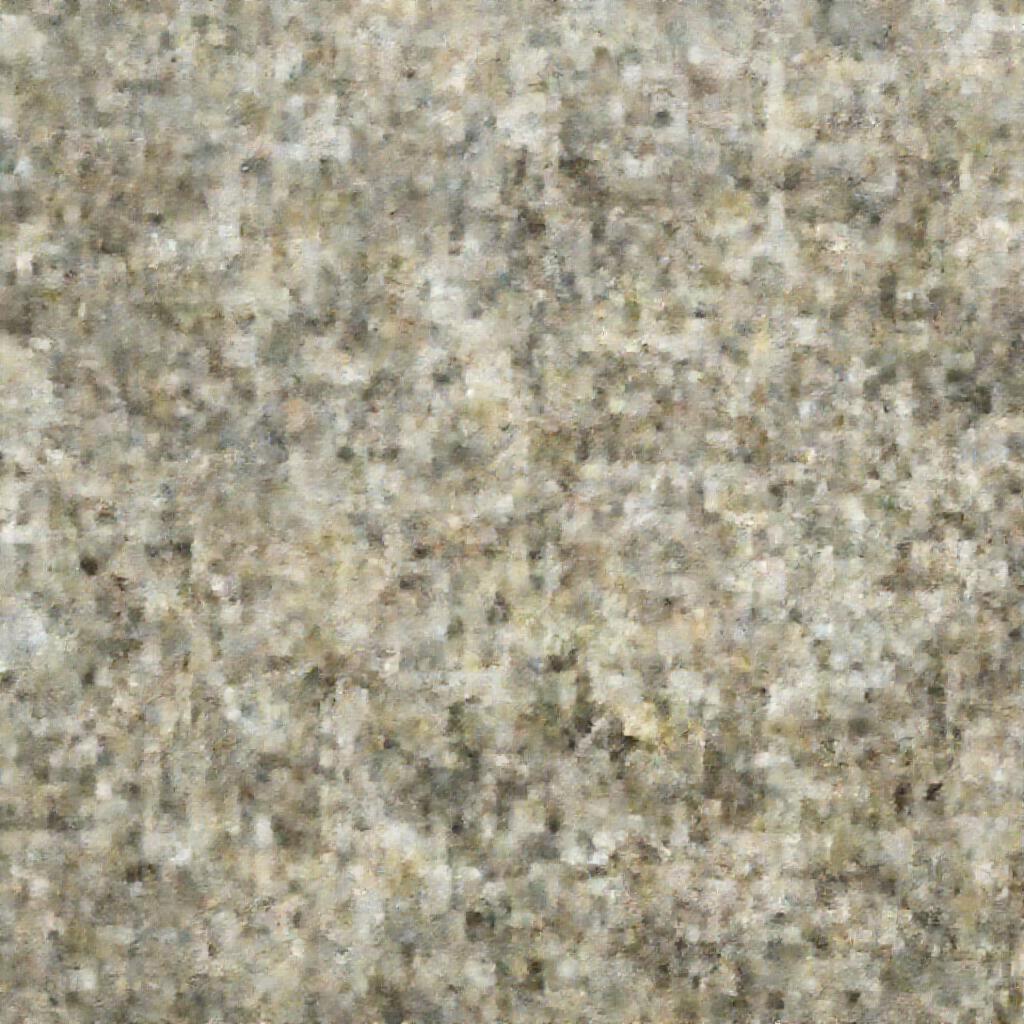} &
        \includegraphics[width=\imgwidth]{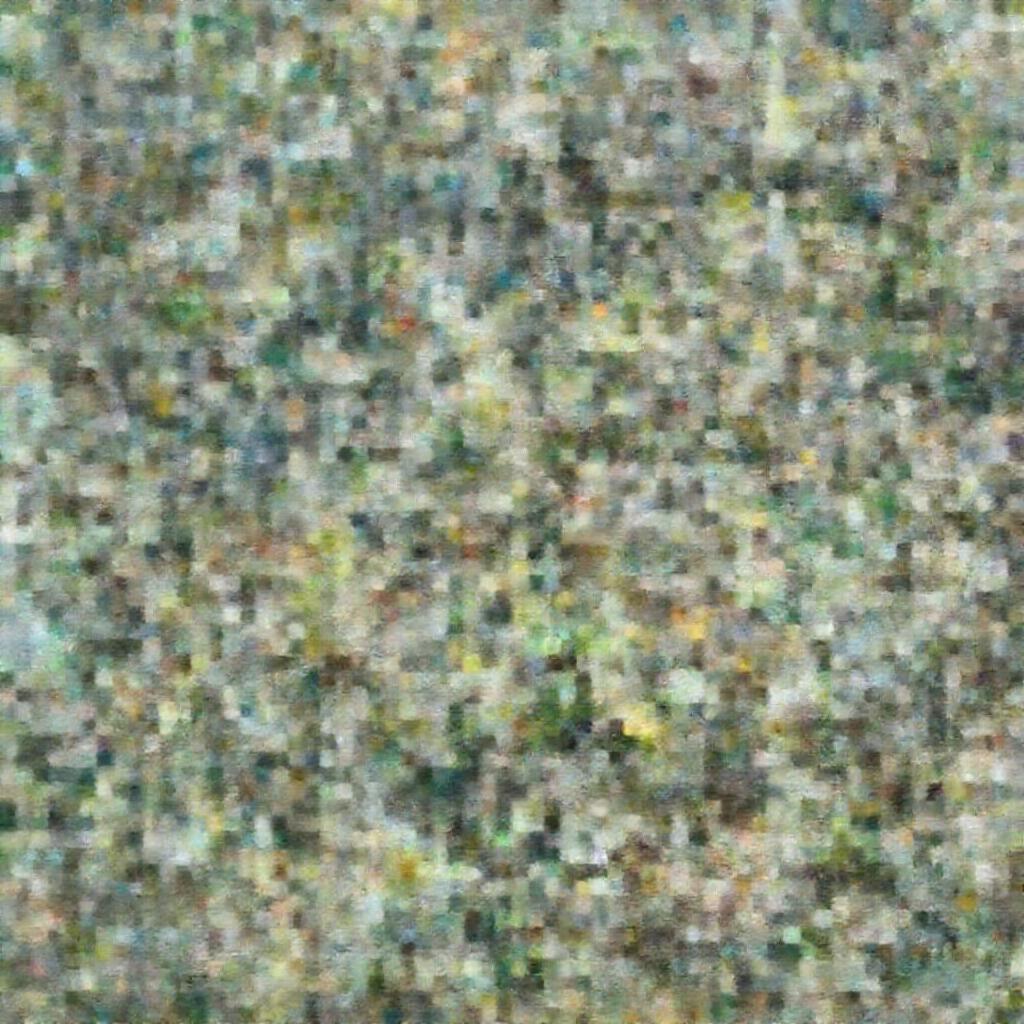} &
        \includegraphics[width=\imgwidth]{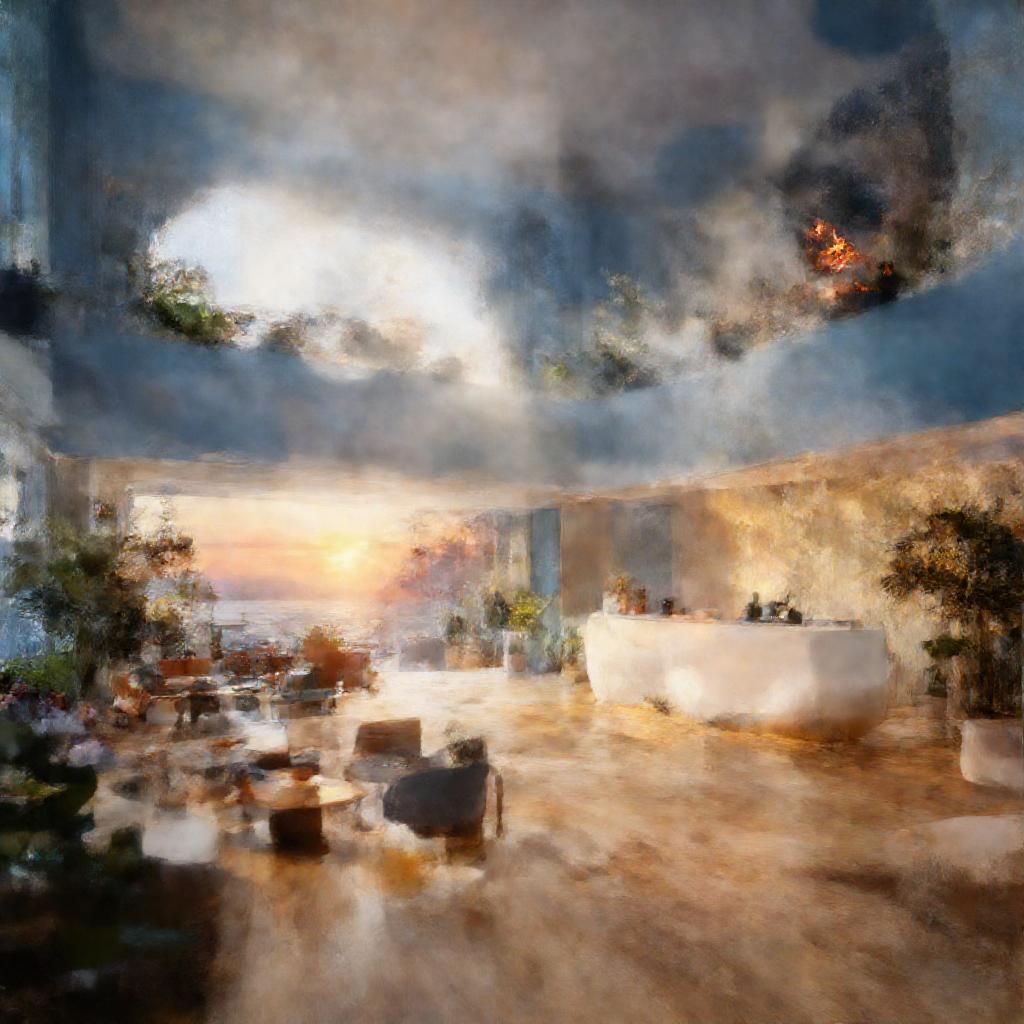} &
        \includegraphics[width=\imgwidth]{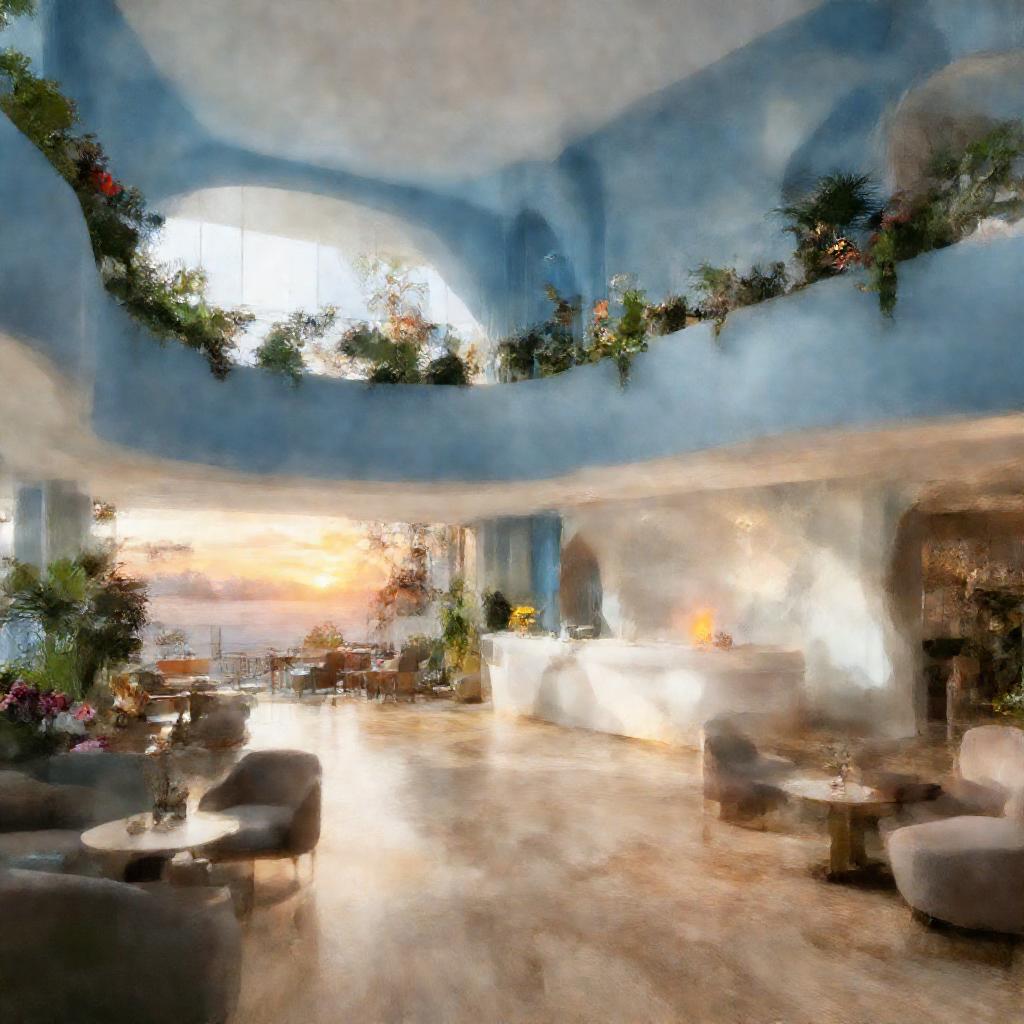} &
        \includegraphics[width=\imgwidth]{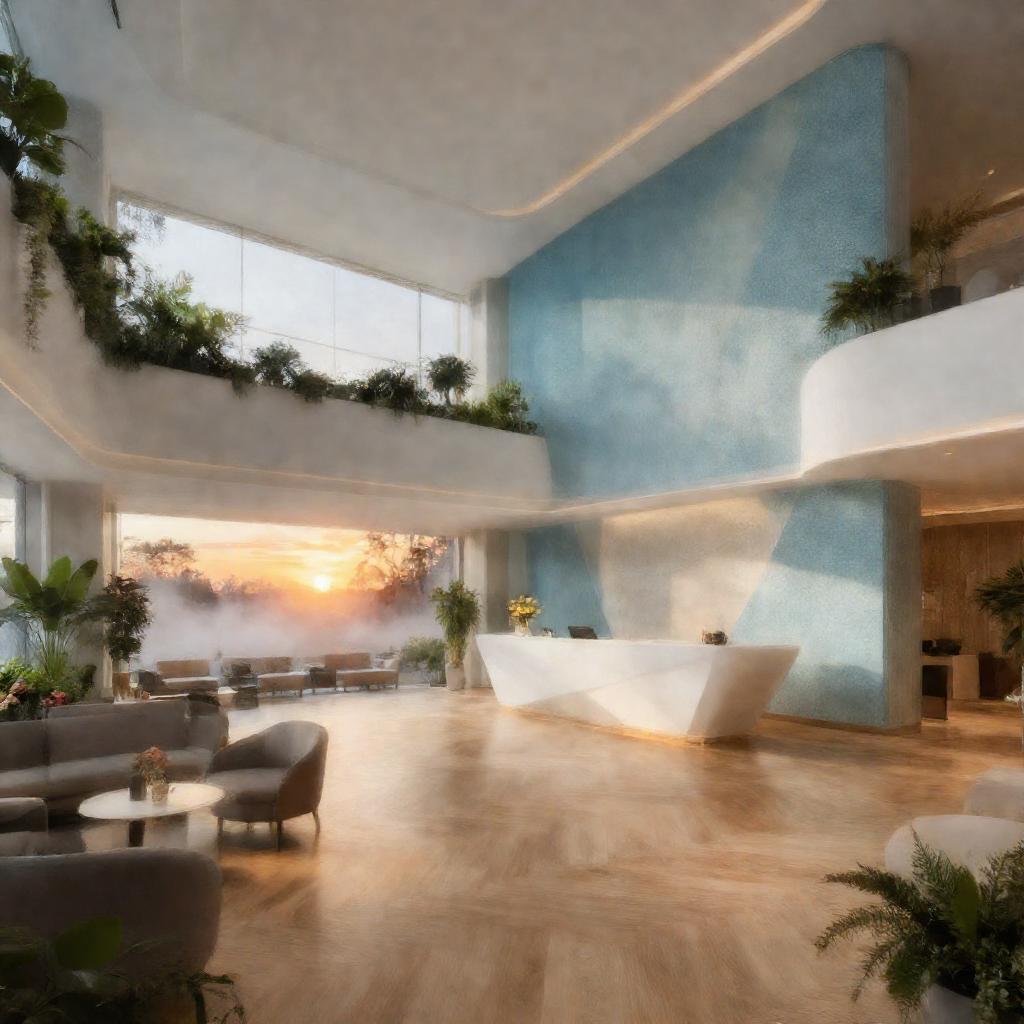} &
        \includegraphics[width=\imgwidth]{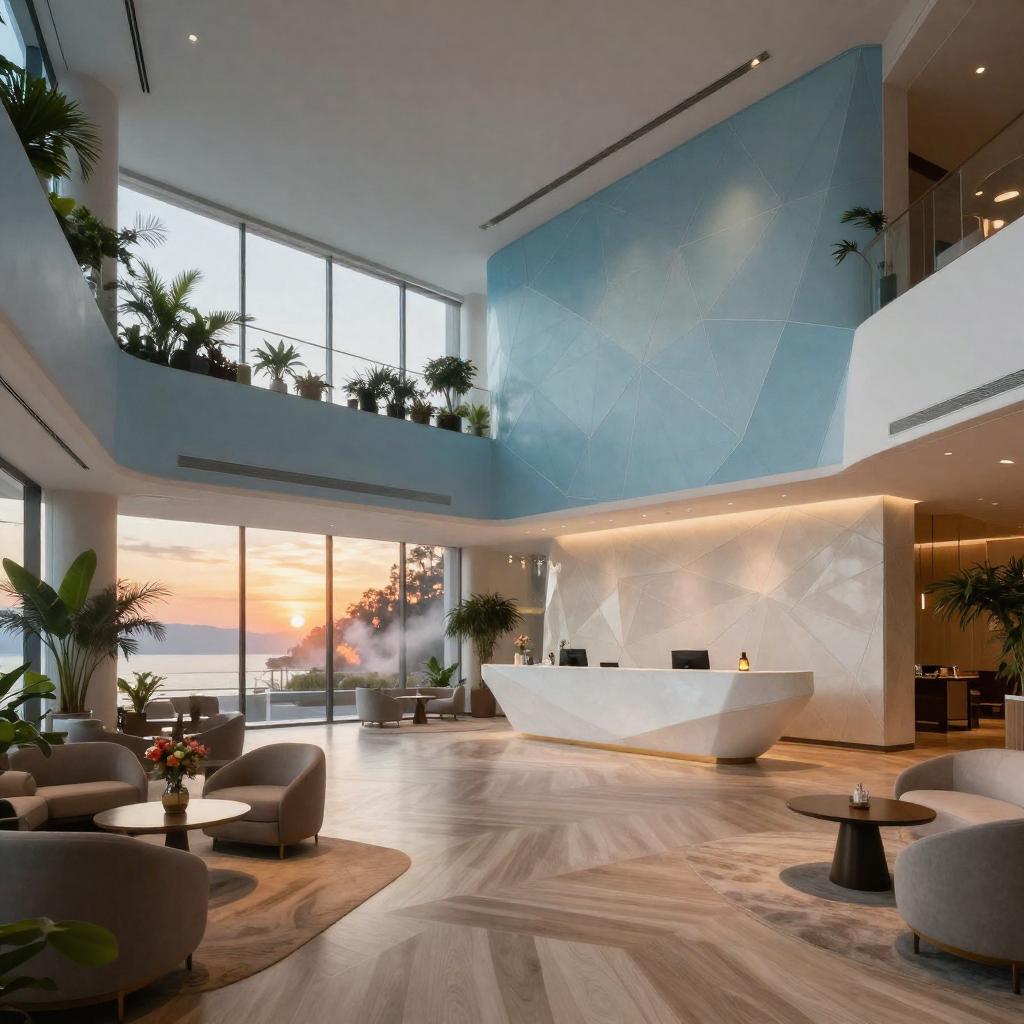} &
        \includegraphics[width=\imgwidth]{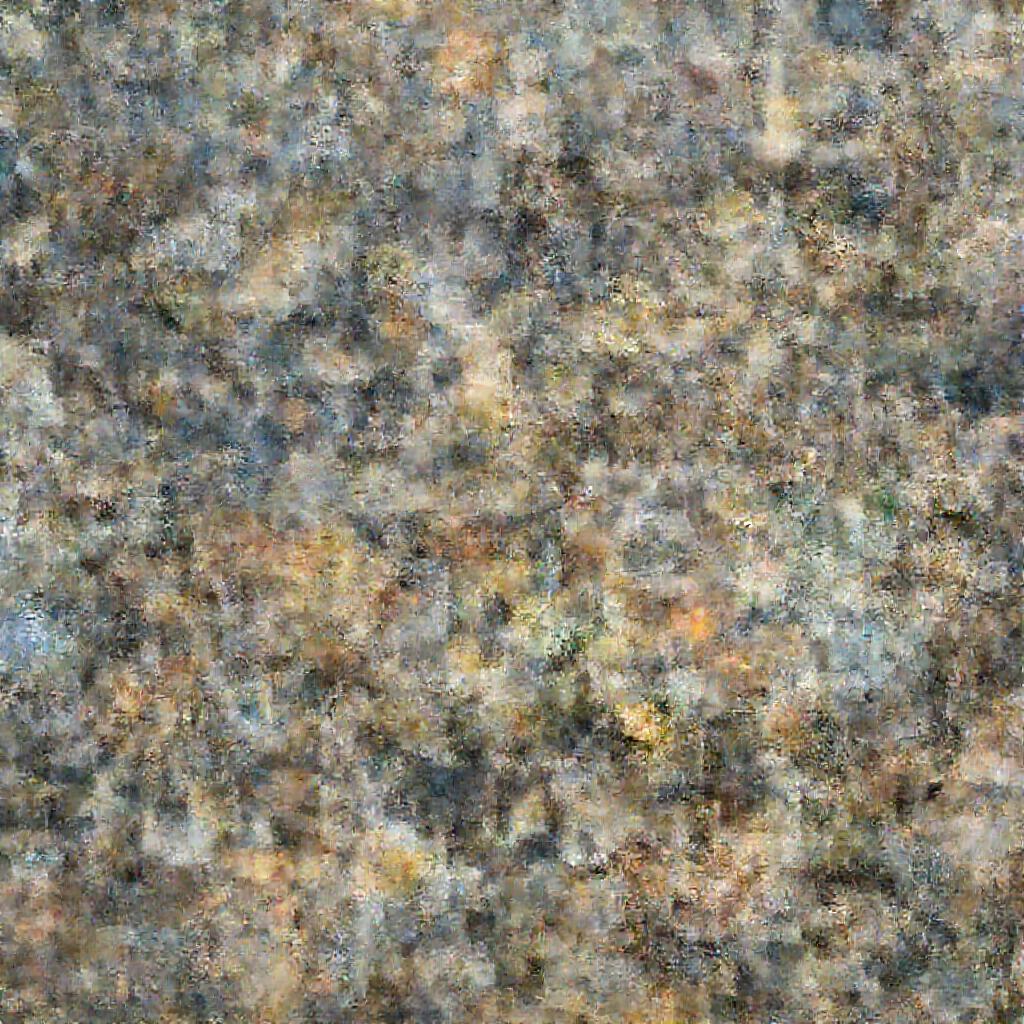} &
        \includegraphics[width=\imgwidth]{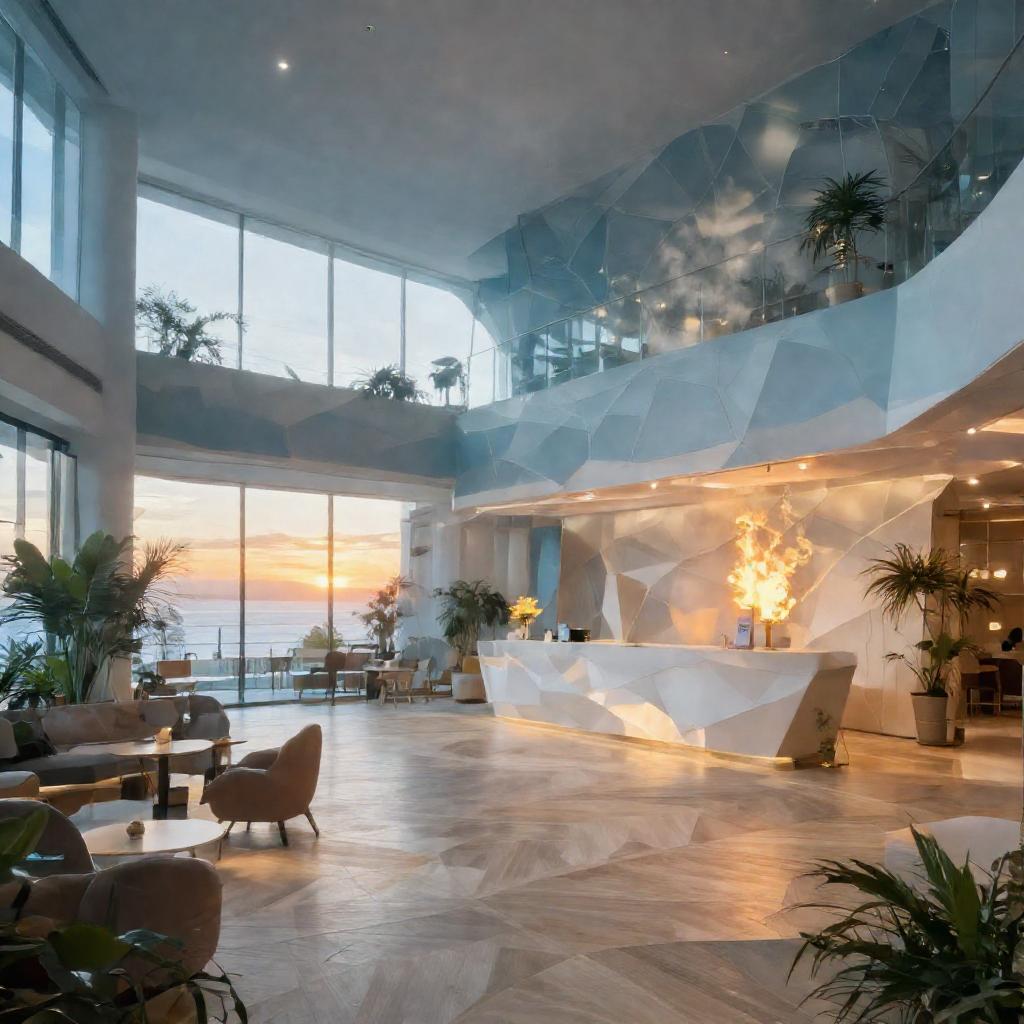} \\

        % --- Row 3 Prompt ---
        \multicolumn{9}{c}{\parbox{0.95\textwidth}{ \scriptsize Model: Z-Image. \textit{Prompt: The parametric hotel lobby is a sleek and modern space with plenty of natural light. The lobby is spacious and open with...}}} \\

    \end{tabular}
\vspace{-3mm}
    \caption{Visual Comparison on Text-to-Image Models. Visual samples from Flux-Dev, Flux-Schnell, and Z-Image generated by AdaTSQ and baselines. AdaTSQ preserves fine-grained details and semantic alignment.}
    \vspace{-2mm}
    \label{fig:visual-compare}
\end{figure*}
\begin{figure*}[t!]
    \centering
    \includegraphics[width=\textwidth]{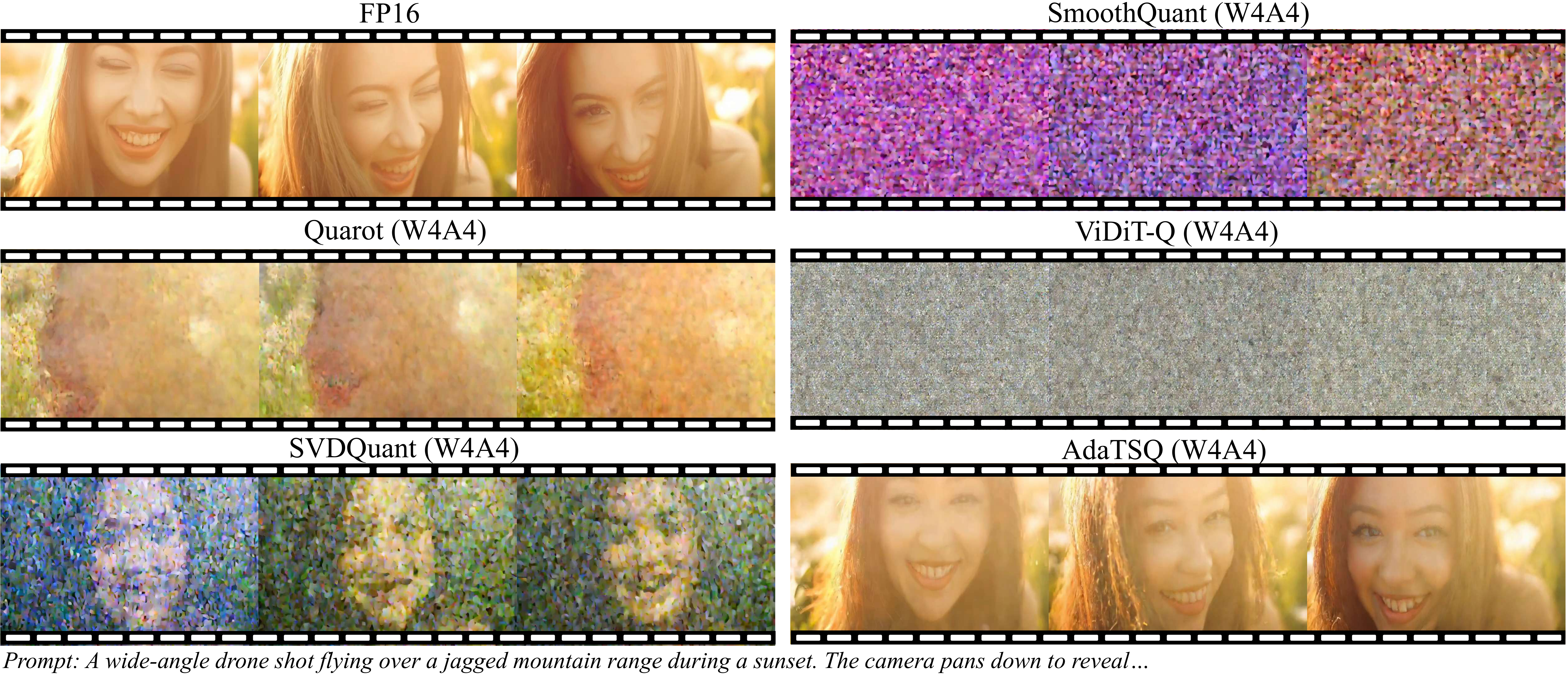}
    \vspace{-6mm}
    \caption{Visual Comparison on Text-to-Video Model. Visual samples from Wan2.1-1.3B generated by AdaTSQ and baselines.}
    \label{fig:video-compare}
    \vspace{-4mm}
\end{figure*}
\vspace{-1mm}
\section{Experiments}
\label{sec:experiments}
\vspace{-2mm}
\subsection{Experimental Setup}
\vspace{-1.5mm}
\textbf{Models \& Datasets.} We evaluate AdaTSQ on four state-of-the-art DiT architectures, covering diverse modalities and sampling steps: (1) Flux-Dev (50-step) and (2) Flux-Schnell (4-step) (3) Z-Image (10-step) for text-to-image generation; (4) Wan2.1-1.3B (25-step) for text-to-video generation. For quantitative evaluation, we employ Geneval for text-to-image models, measuring semantic alignment and fidelity. For text-to-video models, we use VBench to assess temporal consistency and visual quality. 

\vspace{-1.5mm}
\textbf{Baselines.} We compare AdaTSQ against a comprehensive suite of quantization methods, including: (1) \textbf{DiT-specific approaches} such as SVDQuant \cite{li2024svdquant} (current SOTA) and ViDiT-Q \cite{zhao2024viditq}; (2) \textbf{LLM-adapted techniques} like QuaRot \cite{ashkboos2024quarot} and SmoothQuant \cite{xiao2023smoothquant}, which represent advanced static PTQ; and (3) \textbf{earlier diffusion quantization methods} including Q-DiT \cite{chen2024QDiT}. All baselines are evaluated under their optimal settings.

\begin{table*}[t]
    \centering
    \scriptsize
    \caption{Ablation Study on Flux-Dev (W3A3). We evaluate the contribution of Pareto-aware Allocation (``Pareto'') and Fisher-Guided Calibration (``Fisher Calib''). ``Avg Calib'' denotes standard GPTQ with uniform temporal weighting.}
    \label{tab:ablation_study}
    \vspace{-2mm}
    \setlength{\tabcolsep}{3mm}
    \resizebox{\textwidth}{!}{%
    \begin{tabular}{l|ccccccc}
    
\hline
\toprule[0.15em]
\rowcolor{colorhead} Method & Single Object & Position & Counting & Two Object & Colors & Color Attribute & Total \\
\midrule[0.15em]
1. Baseline (MinMax W)              & 0.5594 & 0.0175 & 0.2688 & 0.0985 & 0.3484 & 0.0500 & 0.2238 \\
2. + Pareto Alloc                   & 0.8562 & 0.0725 & 0.5031 & 0.3409 & 0.5771 & 0.1675 & 0.4196 \\
3. + Fisher Calib                   & 0.8125 & 0.0385 & 0.4719 & 0.2803 & 0.4920 & 0.4191 & 0.4190 \\
4. + Pareto Alloc + Avg Calib       & 0.8812 & 0.0750 & 0.5717 & 0.5197 & 0.5214 & 0.2272 & 0.4660 \\
\textbf{5. AdaTSQ}& \textbf{0.9562} & \textbf{0.0825} & \textbf{0.6281} & \textbf{0.5480} & \textbf{0.6559} & \textbf{0.2900} & \textbf{0.5270} \\
\bottomrule[0.15em]

    \end{tabular}
    } % resize box
    \vspace{-2mm}
\end{table*}

\vspace{-2mm}
\subsection{Main Results}
% We present key results in Tables \ref{tab:geneval_results} and \ref{tab:vbench_results}, with full data and additional visual comparisons in Appendix C and D.
\vspace{-1mm}
\subsubsection{Quantitative Comparison}
\vspace{-1mm}
\textbf{Text-to-Image Generation.} Figure \ref{fig:visual-compare} presents qualitative results across three text-to-image models, including Flux-dev, Flux-Schnell and Z-Image. Primitive methods like SmoothQuant and Q-DiT fail completely at W4A4, generating mostly pure noise. While QuaRot and ViDiT-Q capture basic outlines at W4A4, they lack fine-grained details. SVDQuant achieves high fidelity at W4A4 but suffers from significant quality degradation when pushed to W3A3. In contrast, AdaTSQ consistently produces high-quality images that are virtually indistinguishable from the FP16 across both W4A4 and the challenging W3A3 settings.

\vspace{-2mm}
\textbf{Text-to-Video Generation.} In Table \ref{tab:vbench_results}, on Wan2.1, aggressive W4A4 quantization causes catastrophic failure in several baselines (scores dropping to 0). Notably, SVDQuant struggles with video generation at w4a4. Conversely, AdaTSQ demonstrates superior generalization, achieving faithful results across all metrics, effectively handling the temporal complexity of video generation.

% \begin{figure*}[t!]
%     \centering
%     \includegraphics[width=\textwidth]{ICML26/figures/visual-compare/ablation/ablation-compare.pdf}
%     \vspace{-7.5mm}
%     \caption{Visual Comparison on ablation study, which are genarated by Flux-Dev model.}
%     \label{fig:ablation-compare}
%     \vspace{-6mm}
% \end{figure*}
\begin{figure*}[t]
    \centering
    \setlength{\tabcolsep}{1pt} 
    % 定义图片宽度，稍微设大一点点(0.16)充分利用版面，0.15*6=0.9，0.16*6=0.96
    \newcommand{\imgwidth}{0.16\textwidth} 

    % 只有图片行使用 0.5 的行距，避免影响文字行
    \renewcommand{\arraystretch}{0.5} 

    \begin{tabular}{cccccc}
        % --- Header Row (Method Names) ---
        % 注意：这里行尾必须加 \\，并建议加 \vspace 稍微拉开文字和图片的距离
        \scriptsize FP16 & \scriptsize Baseline & \scriptsize + Pareto Alloc & \scriptsize + Fisher Calib  & \scriptsize + Pareto Alloc \& Avg Calib & \scriptsize AdaTSQ \\

        % --- Row 1 Images ---
        \includegraphics[width=\imgwidth]{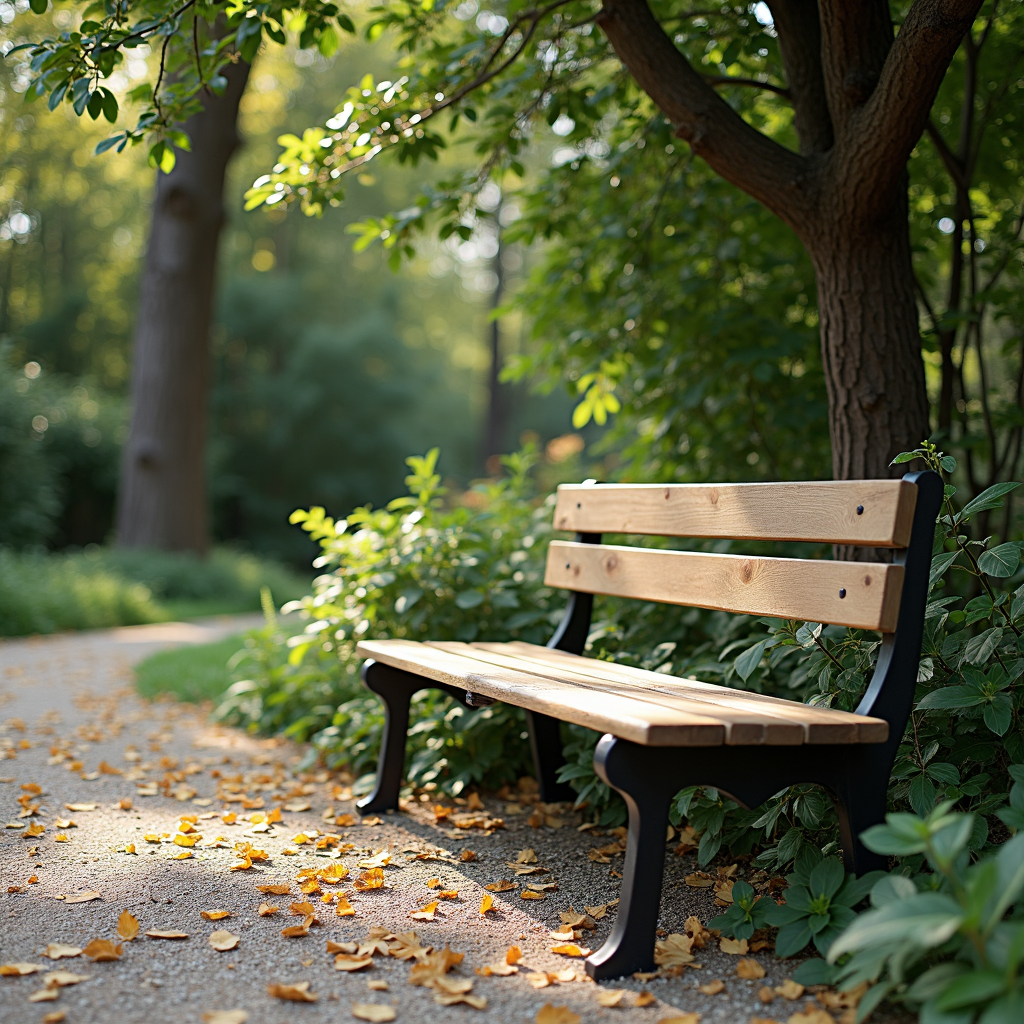} &
        \includegraphics[width=\imgwidth]{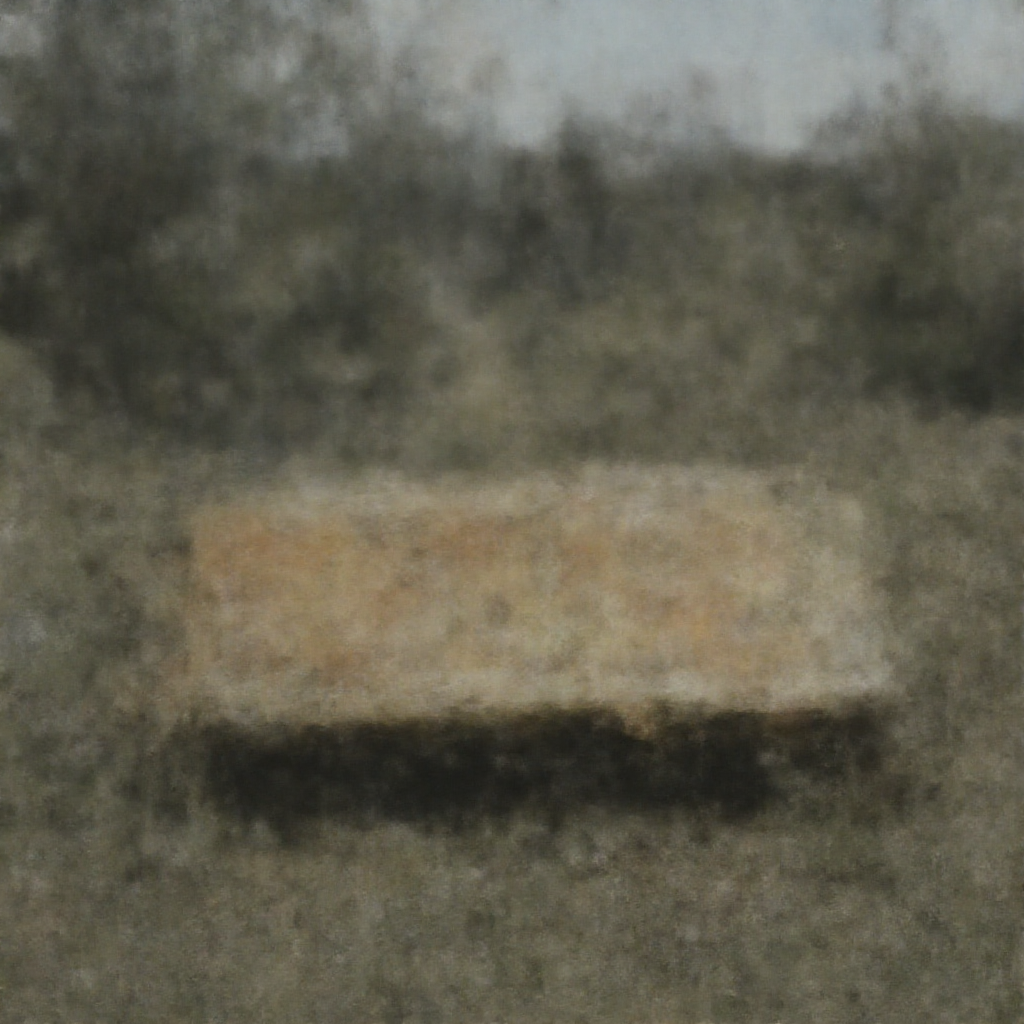} &
        \includegraphics[width=\imgwidth]{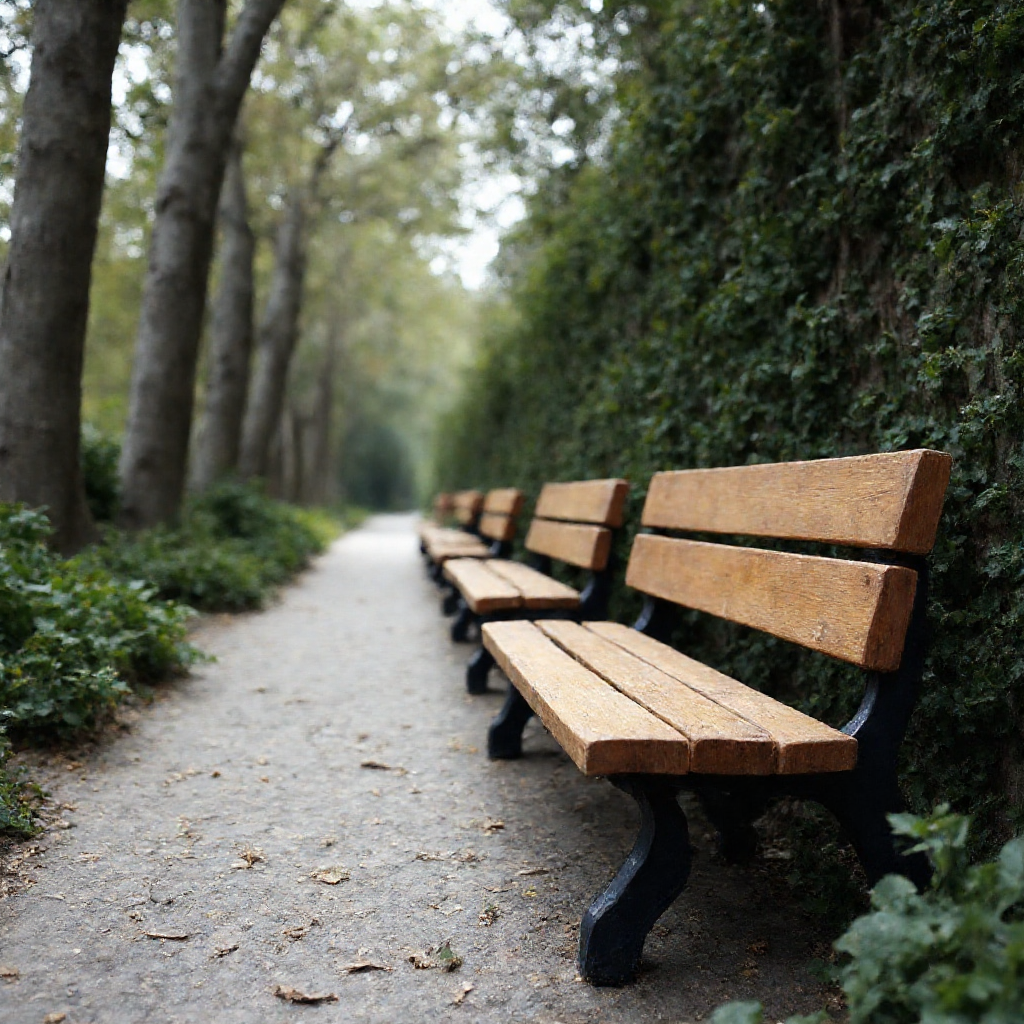} &
        \includegraphics[width=\imgwidth]{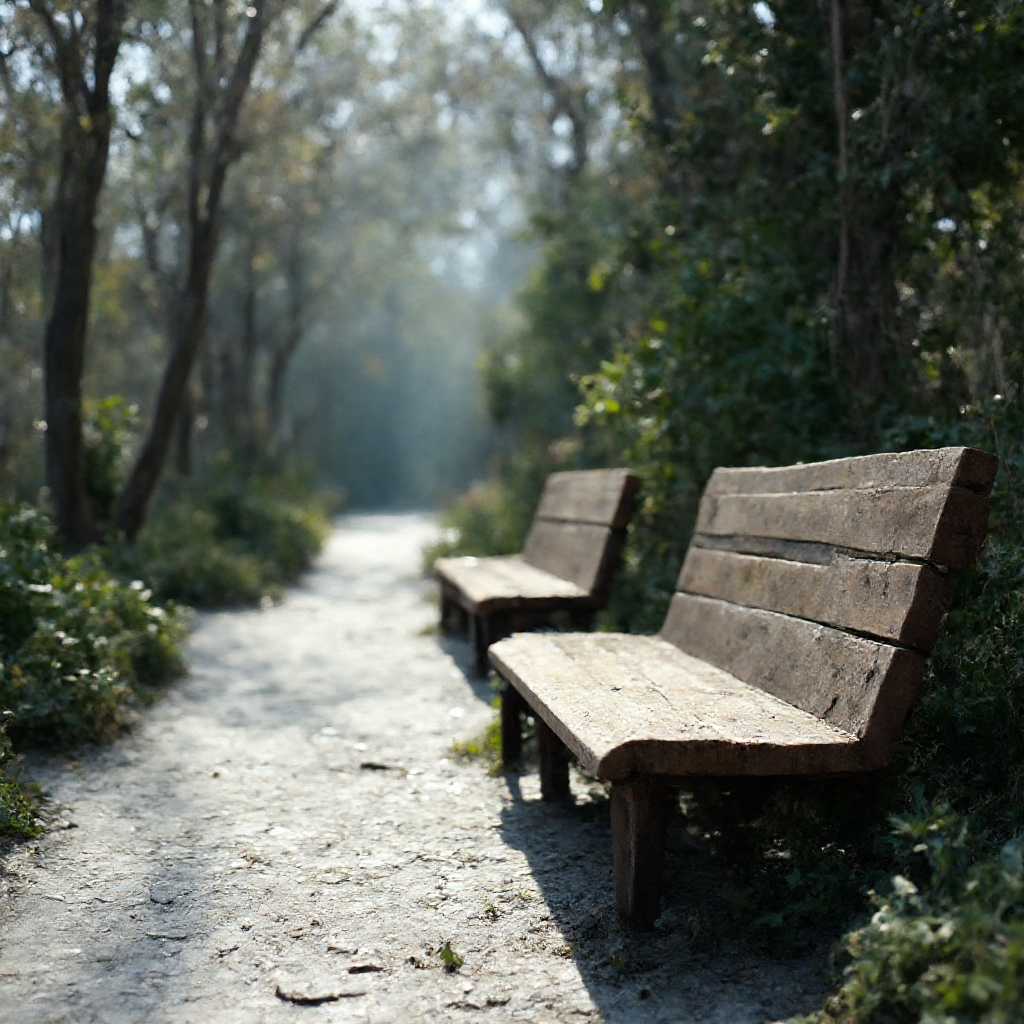} &
        \includegraphics[width=\imgwidth]{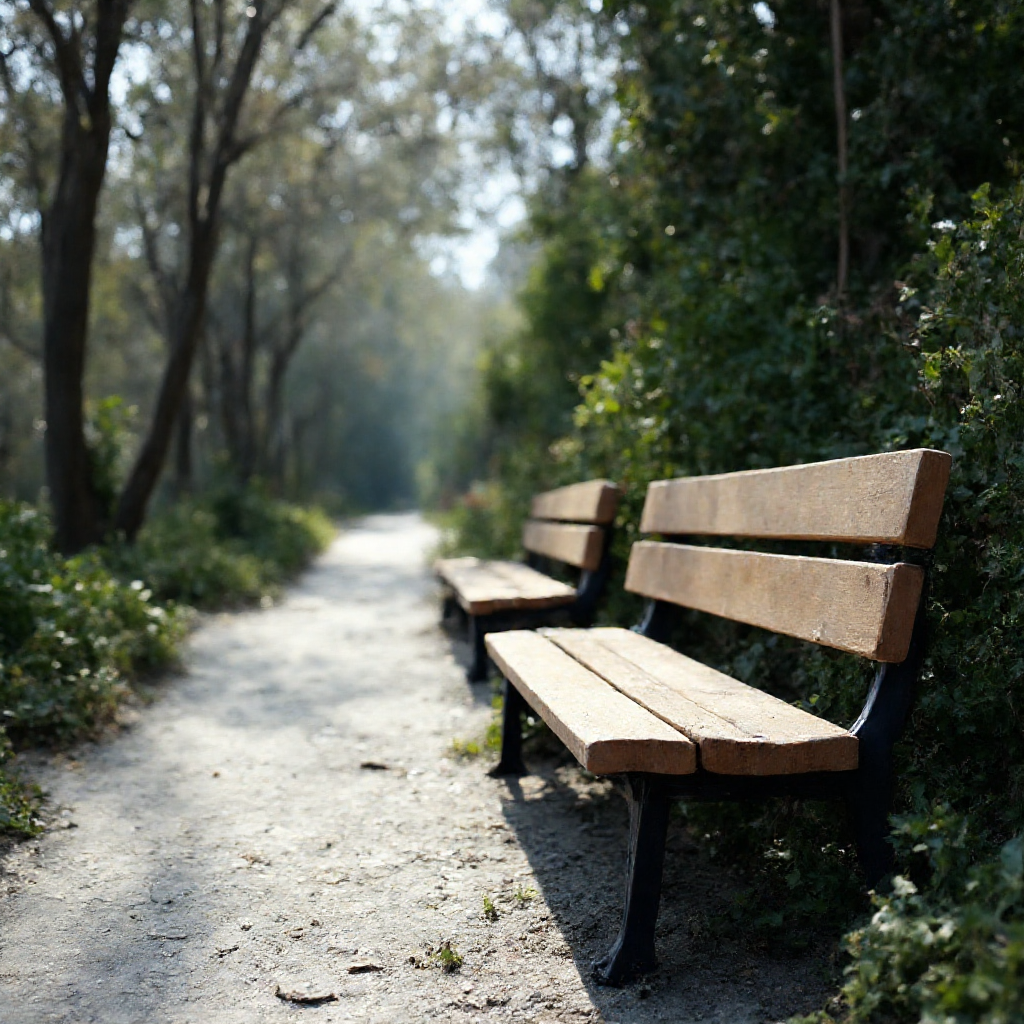} &
        \includegraphics[width=\imgwidth]{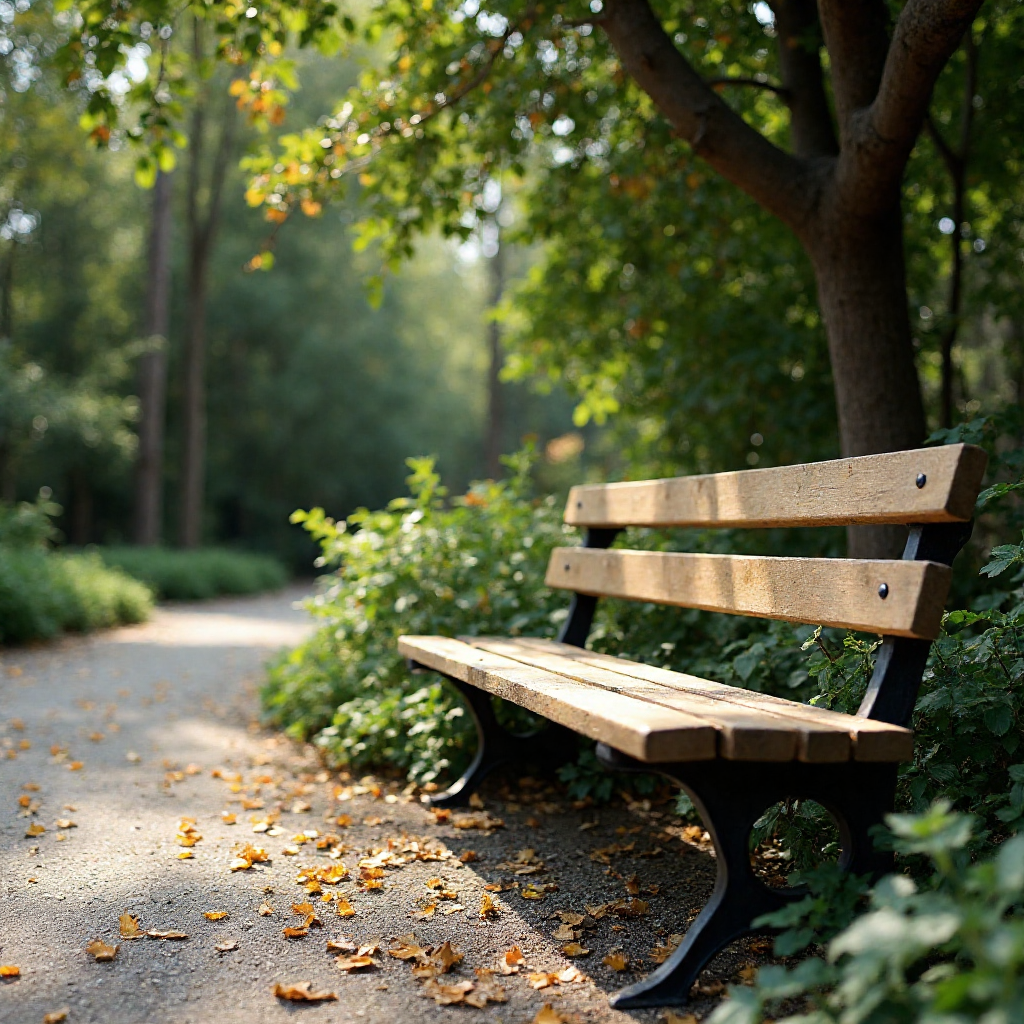} \\

        % --- Prompt Row ---
        \multicolumn{6}{l}{\parbox{0.95\textwidth}{\centering \scriptsize \textit{Prompt: A photo of a bench.}}} \\
    \end{tabular}
    \vspace{-2mm}
    % 修正拼写错误 genarated -> generated
    % 修正 Label 以匹配正文引用
    \caption{Visual comparison of the ablation study, which is generated by Flux-Dev model under W3A3 setting.}
    \vspace{-3mm}
    \label{fig:ablation-compare} 
\end{figure*}
\vspace{-3mm}
\subsubsection{Visual Comparison} 
\vspace{-2mm}
\textbf{Text-to-Image Generation.} Figure \ref{fig:visual-compare} presents qualitative results across three text-to-image models. Primitive methods like SmoothQuant and Q-DiT fail completely at W4A4, generating mostly pure noise. While QuaRot and ViDiT-Q capture basic outlines at W4A4, they lack fine-grained details. SVDQuant achieves high fidelity at W4A4 but suffers from significant quality degradation when pushed to W3A3. In contrast, AdaTSQ consistently produces high-quality images that are virtually indistinguishable from the FP16 across both W4A4 and the challenging W3A3 settings.

\vspace{-1mm}
\textbf{Text-to-Video Genreation.} Figure \ref{fig:video-compare} visualizes the video generation results of AdaTSQ and baselines. Note that Q-DiT results are omitted as they consist entirely of noise. Under the aggressive W4A4 quantization, both SmoothQuant and ViDiT-Q fail to generate meaningful content, producing only noise; this observation suggests that their scaling-based operations might be ill-suited for the specific characteristics of the Wan model architecture. While QuaRot and SVDQuant manage to preserve basic structural outlines, they still struggle to reconstruct fine-grained details, leading to visually flat outputs. In contrast, AdaTSQ generates videos with remarkably high quality, maintaining both temporal consistency and visual fidelity, thereby mitigating the jitter and distortion observed in low-bit video synthesis.

\subsection{Ablation Studies} \label{sec:ablation}
\vspace{-1mm}
To validate the contribution of each component within AdaTSQ, we conduct a comprehensive ablation study on Flux-Dev under the W3A3 setting. The quantitative results are summarized in Table \ref{tab:ablation_study}, and qualitative comparisons are visualized in Figure \ref{fig:ablation-compare}. The naive baseline (Row 1, using minmax for weight quantization and naive rotation for activation quantization) suffers from severe degradation, producing images with collapsed structures.

\vspace{-1mm}
\textbf{Effect of Pareto-aware Allocation.} Introducing Pareto-aware allocation (Row 2, employing a timestep-wise mixed-precision strategy of 3/4/8-bit) significantly boosts image clarity compared to the baseline by assigning higher precision to sensitive timesteps. However, as shown in Figure \ref{fig:ablation-compare}, semantic alignment remains chaotic, often generating content unrelated to the text prompt.

\vspace{-1mm}
\textbf{Effect of Fisher-Guided Calibration.} Applying Fisher-guided calibration alone (Row 3) partially corrects semantic errors but still exhibits noticeable deviations. This confirms that risk-aware weight optimization—leveraging Fisher information to identify and preserve critical parameters—is crucial for aligning the generation path, but its effectiveness is limited when constrained by fixed bit-widths.

\vspace{-1mm}
\textbf{Synergy and Final Performance.} Combining Pareto allocation with standard average calibration (Row 4) brings structural clarity close to our final method's result, yet semantic flaws persist due to the lack of sensitivity-aware adjustment (the prompt requires only one bench, but there exists two). The optimal result is consistently achieved by our full method (Row 5). As visualized, AdaTSQ not only maintains high visual clarity but also perfectly resolves semantic alignment constraints, demonstrating that the synergy between dynamic bit allocation and risk-aware calibration is essential for achieving robust generation performance under the extreme W3A3 setting.

\begin{table}[t]
    \centering
    \caption{Theoretical Efficiency on Flux-Dev (12B). Comparison of average bit-width, memory footprint (Model Size), and theoretical computational cost (FLOPs). AdaTSQ achieves $>$5$\times$ reduction across all metrics.}
    \label{tab:flux_efficiency}
    \vspace{-2mm}
    \setlength{\tabcolsep}{3.5mm}
    \resizebox{\columnwidth}{!}{
    \begin{tabular}{c|ccc}
    \toprule
    \rowcolor{colorhead}
    Metric & FP16 & AdaTSQ & Reduction \\
    \midrule
    Avg. Bit-width & W16A16 & W3A3.1 & \textbf{5.16$\times$} \\
    Model Size & $\sim$24.0 GB & $\sim$4.50 GB & \textbf{5.33$\times$} \\
    Norm. FLOPs & 1.00$\times$ & 0.19$\times$ & \textbf{5.16$\times$} \\
    \bottomrule
    \end{tabular}
    }
    \vspace{-5mm}
\end{table}
\subsection{Efficiency Analysis}
\label{sec:efficiency}
\vspace{-2mm}
\textbf{Search \& Inference Efficiency.} A key advantage of AdaTSQ is its minimal overhead. The Pareto-aware beam search is extremely efficient; on a single A100-80GB GPU, finding the optimal mixed-precision policy for the 50-step Flux-Dev model requires only $\sim$4 minutes. This negligible one-off cost unlocks substantial theoretical gains, as summarized in Table \ref{tab:flux_efficiency}. Based on the searched policy, the bit-width distribution converges to approximately 80\% 3-bit, 10\% 4-bit, and 10\% 8-bit. This results in an effective average bit-width of 3.1 bits, translating to a theoretical \textbf{5.16$\times$ reduction} in computational FLOPs and \textbf{5.33$\times$} in model size and compared to the FP16 baseline.

\vspace{-3mm}
\section{Conclusion}
\label{sec:conclusions}
\vspace{-2mm}
We presented \textbf{AdaTSQ}, a novel PTQ framework that systematically exploits the temporal heterogeneity of Diffusion Transformers. By quantizing activation with Pareto-aware timestep-dynamic allocation and introducing Fisher-guided temporal weight calibration, AdaTSQ effectively resolves the conflict between static quantization parameters and dynamic activation distributions. Extensive experiments demonstrate that our method achieves state-of-the-art performance, enabling perceptually  W4A4 quantization across diverse image and video models. Most notably, we unlock viable W3A3 generation on text-to-image models for the first time, revealing the inherent robustness of image genaration models to low-bit compression.
\newpage
\twocolumn
\bibliography{ICML26/refs}
\bibliographystyle{icml2026}
\end{document}